\documentclass[10pt, journal, final]{IEEEtran}

\usepackage{tikz}
\usepackage{verbatim}
\usetikzlibrary{arrows}
\usetikzlibrary{shapes,snakes}
\usetikzlibrary{decorations.pathmorphing} 
\usetikzlibrary{fit}					
\usetikzlibrary{backgrounds}	

\usepackage{balance}

\newcommand{\eg}{\emph{e.g.}}
\newcommand{\ie}{\emph{i.e.}}

\newcommand{\abs}[1]{\left|#1\right|}

\usepackage{amsmath}
\usepackage{amsfonts}
\usepackage{amssymb}
\usepackage{graphicx}
\usepackage{comment}
\usepackage{color}

\usepackage[notheorems]{rr-math}

\usepackage{tikz}
\usepackage{verbatim}
\usetikzlibrary{arrows}

\makeatletter
\global\let\tikz@ensure@dollar@catcode=\relax
\makeatother

\usepackage{tabularx}
\newcolumntype{H}{>{\setbox0=\hbox\bgroup}c<{\egroup}@{}}

\usepackage{ifpdf}
\usepackage{xcolor}
\usepackage{graphicx,epsfig}
\usepackage{graphics,dcolumn,bm,epic,eepic,float}
\usepackage{amssymb,amsmath,multirow,rotate,color}
\usepackage{subfigure}
\usepackage{epstopdf}
\usepackage{tabularx}
\usepackage{booktabs}

\usepackage{amsmath,amsfonts,amssymb}
\usepackage{multirow}
\usepackage{comment}
\usepackage{rotating}
\usepackage{afterpage}

\usepackage{graphicx,epsfig}
\usepackage{graphics,dcolumn,bm,epic,eepic,float}
\usepackage{amssymb,amsmath,multirow,rotate,color}
\usepackage{subfigure}
\usepackage{epstopdf}
\usepackage{pifont}

\newcommand{\etal}{\emph{et al.}\xspace}

\newcommand{\cmark}{\ding{51}}
\newcommand{\xmark}{\ding{55}}

\definecolor{gray}{RGB}{150,150,150}
\colorlet{TufteRed}{red!80!black}

\definecolor{thelightblue}{RGB}{0,191,255}
\definecolor{theblue}{RGB}{0,0,180}
\colorlet{thered}{TufteRed}

\usepackage{paralist}

\usepackage{url}
\usepackage{amssymb}
\usepackage{graphicx}
\usepackage{helvet}
\usepackage{courier}
\usepackage{subfigure}
\usepackage{amsmath}
\usepackage{multirow}
\usepackage{comment}
\usepackage{amssymb}
\usepackage{multirow}
\usepackage{color}

\usepackage{relsize}
\usepackage{algorithm}
\usepackage{algorithmicx}
\usepackage{algpseudocode}

\algblockdefx[parallel]{ParFor}{EndPar}[1][]{$\textbf{parallel for}$ #1 $\textbf{do}$}{$\textbf{end parallel}$}
\algrenewcommand{\alglinenumber}[1]{\fontsize{6.5}{7}\selectfont#1}
\algtext*{EndPar}

\algblockdefx[parallel]{parfor}{endpar}[1][]{$\textbf{parallel for}$ #1 $\textbf{do}$}{$\textbf{end parallel}$}
\algrenewcommand{\alglinenumber}[1]{\scriptsize#1:}

\usepackage{nicefrac}

\algnotext{EndFor}
\algnotext{EndIf}
\algnotext{EndProcedure}

\ifpdf
\usepackage{epstopdf}
\fi
\graphicspath{{./}{./figures/}}

\usepackage{graphics}
\usepackage{rotating}
\usepackage{wrapfig}
\usepackage{setspace}
\usepackage{xspace}

\usepackage{amsmath}
\usepackage{amssymb}
\usepackage{textcase}
\usepackage{soul}
\usepackage{dashrule}

\usepackage{url}

\newcommand{\be}{\begin{equation}}
\newcommand{\ee}{\end{equation}}
\newcommand{\bea}{\begin{eqnarray}}
\newcommand{\eea}{\end{eqnarray}}
\newcommand{\bit}{\begin{itemize}}
\newcommand{\eit}{\end{itemize}}

\newcommand{\cm}[1]{}
\newcommand{\jt}[1]{}

\usepackage{bold-extra}

\usepackage{amsthm}
\makeatletter
\def\th@plain{
\thm@notefont{}
\itshape 
}
\def\th@definition{
\thm@notefont{}
\normalfont 
}
\makeatother

\newtheoremstyle{probstyle}
{1.1mm}
{1.1mm}
{\itshape}
{}
{\bfseries}
{:}
{ }
{}
\theoremstyle{probstyle}

\newtheoremstyle{lemstyle}
{0mm}
{0mm}
{\itshape}
{}
{\bfseries\sc}
{. }
{ }
{}
\theoremstyle{lemstyle}
\newtheorem{Claim}{Claim}[]
\newtheorem{Lemma}{Lemma}[]  

\usepackage{bold-extra}
\newtheoremstyle{mystyle}
{1.5mm}
{1.5mm}
{\itshape}
{}
{\bfseries\scshape}
{ \sc}
{ }
{}
\theoremstyle{mystyle}
\newtheorem{Definition}{Definition}[]

\newtheoremstyle{newstyle}
{1.0mm}
{1.0mm}
{\itshape}
{}
{\bfseries}
{}
{ }
{}
\theoremstyle{newstyle}

\newtheoremstyle{nonumberplain}
{-0.2}
{}
{\normalfont\upshape}
{1mm}
{\normalfont\small\sffamily\sc}
{:}
{ }
{}
\theoremstyle{nonumberplain}

\definecolor{red}{RGB}{0,0,0}

\renewcommand{\Pr}{\ensuremath{\mathbb{P}}}

\ifCLASSINFOpdf
\else
\fi

\usepackage{xfrac}    
\usepackage{nicefrac} 
\usepackage{etoolbox}

\usepackage{graphicx}
\usepackage{amssymb}
\usepackage{amsmath}
\usepackage{multirow}
\usepackage{rotate}
\usepackage{subfigure}
\usepackage{epstopdf}
\usepackage{tabularx}
\usepackage{booktabs}
\usepackage{paralist}
\usepackage{graphicx}
\usepackage{ifpdf}
\usepackage{mathdots}
\usepackage{lscape}
\usepackage{setspace}
\usepackage{relsize}
\usepackage{amsfonts}

\usepackage{pifont}

\usepackage[    
protrusion=true, 
kerning=true,
spacing=true,
letterspace=40,
shrink=40,      
factor=1000
]    
{microtype} 

\ifCLASSOPTIONcompsoc
\usepackage[nocompress]{cite}
\else
\usepackage{cite}
\fi

\begin{document}

\title{Dynamic Node Embeddings from Edge Streams}

\author{John Boaz Lee, Giang Nguyen, Ryan A. Rossi, Nesreen K. Ahmed, Eunyee Koh, Sungchul Kim
\thanks{J. Lee and G. Nguyen are with WPI}
\thanks{R.~A. Rossi is with Adobe Research}
\thanks{N.~K. Ahmed is with Intel Labs}
\thanks{E. Koh and S. Kim are with Adobe Research}
\thanks{}
\thanks{\!\!\!\!\!\!\copyright 2020 IEEE.  Personal use of this material is permitted.  Permission from IEEE must be obtained for all other uses, in any current or future media, including reprinting/republishing this material for advertising or promotional purposes, creating new collective works, for resale or redistribution to servers or lists, or reuse of any copyrighted component of this work in other works.}
}

\markboth{IEEE Transactions on Emerging Topics in Computational Intelligence,~Vol.~1, No.~1, July~2020}
{Lee \MakeLowercase{\textit{et al.}}: Dynamic Node Embeddings from Edge Streams}

\maketitle

\begin{abstract}
Networks evolve continuously over time with the addition, deletion, and changing of links and nodes.
Such temporal networks (or edge streams) consist of a sequence of timestamped edges 
and are seemingly ubiquitous.
Despite the importance of accurately modeling the temporal information, most embedding methods ignore it entirely or approximate the temporal network using a sequence of static snapshot graphs.
In this work, we propose using the notion of \emph{temporal walks} for learning dynamic embeddings from temporal networks.
Temporal walks capture the temporally valid interactions (\eg, flow of information, spread of disease) in the dynamic network in a lossless fashion.
Based on the notion of temporal walks, we describe a general class of embeddings called continuous-time dynamic network embeddings (CTDNEs) that completely avoid the issues and problems that arise when approximating the temporal network as a sequence of static snapshot graphs.
Unlike previous work, CTDNEs learn dynamic node embeddings directly from the temporal network at the finest temporal granularity and thus use only temporally valid information.
As such CTDNEs naturally support online learning of the node embeddings in a streaming real-time fashion.
Finally, the experiments demonstrate the effectiveness of this class of embedding methods that leverage temporal walks as it achieves an average gain in AUC of $11.9\%$ across all methods and graphs. 
\end{abstract}


\IEEEpeerreviewmaketitle

\section{Introduction}
\label{sec:intro}
\IEEEPARstart{D}{ynamic} networks are seemingly ubiquitous in the real-world.
Such networks evolve over time with the addition, deletion, and changing of nodes and links.
The temporal information in these networks is known to be important to accurately model, predict, and understand network data~\cite{watts1998collective,newman2001structure}.
Despite the importance of these dynamics, most previous work on embedding
methods have ignored the temporal information in network data~\cite{deepwalk,node2vec,line,grarep,deepGL,struc2vec,ASNE,ahmed17learning-attr-graphs,ComE,lee17-Deep-Graph-Attention}.

\makeatletter
\global\let\tikz@ensure@dollar@catcode=\relax
\makeatother
\tikzstyle{every node}=[font=\large,line width=1.5pt]
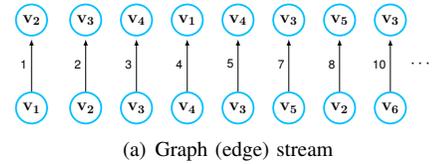
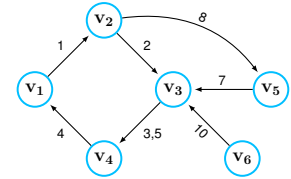
\begin{figure}[t!]
\centering
\begin{center}

\subfigure[Graph (edge) stream]{
\scalebox{0.45}{
\centering
\begin{tikzpicture}[->,>=latex,shorten >=2.4pt,auto,node distance=2.6cm,thick,
main node/.style={circle,draw=thelightblue,fill=white,draw,font=\sffamily\Large\bfseries},
invis node/.style={circle,draw=white,fill=white,draw,font=\sffamily\Large\bfseries,text=black}]

\node[main node] (2) at (0,0) {$\mathbf{v_2}$};
\node[main node] (1) [below of=2] {$\mathbf{v_1}$};

\node[main node] (3) [right of=2] at (-1,0) {$\mathbf{v_3}$};
\node[main node] (22) [below of=3]{$\mathbf{v_2}$};

\node[main node] (4) [right of=3] at (0.5,0) {$\mathbf{v_4}$};
\node[main node] (33) [below of=4]{$\mathbf{v_3}$};

\node[main node] (11) [right of=4] at (2,0) {$\mathbf{v_1}$};
\node[main node] (44) [below of=11]{$\mathbf{v_4}$};

\node[main node] (444) [right of=11] at (3.5,0) {$\mathbf{v_4}$};
\node[main node] (333) [below of=444]{$\mathbf{v_3}$};

\node[main node] (3333) [right of=444] at (5,0) {$\mathbf{v_3}$};
\node[main node] (5) [below of=3333]{$\mathbf{v_5}$};

\node[main node] (55) [right of=3333] at (6.5,0) {$\mathbf{v_5}$};
\node[main node] (222) [below of=55]{$\mathbf{v_2}$};

\node[main node] (33333) [right of=55] at (8,0) {$\mathbf{v_3}$};
\node[main node] (6) [below of=33333]{$\mathbf{v_6}$};

\node[invis node] (0) [right of=33333] at (9.3,0) {$\mathbf{}$};
\node[invis node] (00) [below of=0]{$\mathbf{}$};

\tikzstyle{LabelStyle}=[below=3pt]
\path[every node/.style={font=\sffamily}] 

(1) edge [left] node [anchor=center, left] {1} (2)
(22) edge [left] node [anchor=center, left] {2} (3)
(33) edge [left] node [anchor=center, left] {3} (4)
(44) edge [left] node [anchor=center, left] {4} (11)
(333) edge [left] node [anchor=center, left] {5} (444)
(5) edge [left] node [anchor=center, left] {7} (3333)
(222) edge [left] node [anchor=center, left] {8} (55)
(6) edge [left] node [anchor=center, left] {10} (33333)
(00) edge [thick,line width=0mm,draw=white,left] node [anchor=center, left] {\Large \bf $\cdots$} (0);

\end{tikzpicture}
}
}

\subfigure[Continuous-Time Dynamic Network (CTDN)]{
\scalebox{0.5}{
\centering
\begin{tikzpicture}[->,>=latex,shorten >=2.4pt,auto,node distance=2.6cm,thick,
main node/.style={circle,draw=thelightblue,fill=white,draw,font=\sffamily\Large\bfseries},
white node/.style={circle,draw=white,fill=white,text=white,draw,font=\sffamily\Large\bfseries}]

\node[main node] (3) {$\mathbf{v_2}$};
\node[main node] (1) [below left of=3] {$\mathbf{v_1}$};
\node[main node] (4) [below right of=1] {$\mathbf{v_4}$};
\node[main node] (2) [below right of=3] {$\mathbf{v_3}$};
\node[main node] (5) [right of=2] {$\mathbf{v_5}$};
\node[main node] (6) [below right of=2] {$\mathbf{v_6}$};
\node[white node] (7) [left of=1] {$\mathbf{---}$}; 
\node[white node] (8) [right of=5] {$\mathbf{---}$}; 

\tikzstyle{LabelStyle}=[below=3pt]
\path[every node/.style={font=\sffamily}] 

(1) edge [left] node [above left] {1} (3) 

	(2)  edge [right] node[below right] {3,5} (4)
	
(4) edge [left] node[below left] {4} (1)

	(3) edge[bend left]  node[sloped,anchor=center,above] {8} (5) 
	(5) edge  node[anchor=center,above] {7} (2) 
	(6) edge  node[sloped,anchor=center,below] {10} (2) 

	(3) edge [right] node[above right] {2} (2);
\end{tikzpicture}
}
}
\end{center}

\caption{
Dynamic network. Edges are labeled by time. 
Observe that existing methods that ignore time would consider $v_4 \!\! \longrightarrow \! v_1 \!\! \longrightarrow \! v_2$ a \emph{valid} walk, however, $v_4 \!\! \longrightarrow \! v_1 \!\! \longrightarrow \! v_2$ is clearly \emph{invalid with respect to time} since $v_1 \!\! \longrightarrow \! v_2$ exists in the past with respect to $v_4 \!\! \longrightarrow \! v_1$.
In this work, we propose the notion of \emph{temporal random walks} for embeddings that capture the \emph{true temporally valid} behavior in networks.
In addition, our approach naturally supports learning in \emph{graph streams} where edges arrive continuously over time (\eg, every second/millisecond)
}
\label{fig:info-loss-example}
\end{figure}

In this work, we address the problem of learning dynamic node embeddings directly from edge streams (\ie, \emph{continuous-time dynamic networks}) consisting of a sequence of timestamped edges at the finest temporal granularity for improving the accuracy of predictive models.
We propose \emph{continuous-time dynamic network embeddings} (CTDNE) and describe a general framework for learning such embeddings based on the notion of \emph{temporal random walks} (walks that respect time).
The framework is general with many interchangeable components and can be used in a straightforward fashion for incorporating temporal dependencies into existing node embedding and deep graph models that use random walks.
Most importantly, the CTDNEs are learned from temporal random walks that represent actual \emph{temporally valid sequences} of node interactions 
and thus avoids the issues and information loss that arises when time is ignored~\cite{deepwalk,node2vec,line,grarep,deepGL,struc2vec,ASNE,ahmed17learning-attr-graphs,ComE,lee17-Deep-Graph-Attention} or approximated as a sequence of discrete static snapshot graphs~\cite{rossi2013dbmm-wsdm,hisano2016semi,kamra2017dgdmn,saha2018models,rahman2018dylink2vec} (Figure~\ref{fig:info-discrete-time-model-loss-example}) as done in previous work.
The proposed approach 
(1) obeys the direction of time and 
(2) biases the random walks towards edges (and nodes) that are more recent and more frequent.
The result is a more appropriate time-dependent network representation that captures the important temporal properties of the continuous-time dynamic network at the finest most natural temporal granularity without loss of information while using walks that are temporally valid (as opposed to walks that do not obey time and thus are invalid and noisy as they represent sequences that are impossible with respect to time).
Hence, the framework allows existing embedding methods to be easily adapted for learning more appropriate network representations from continuous-time dynamic networks by ensuring time is respected and avoiding impossible sequences of events.

The proposed framework learns more appropriate dynamic node embeddings directly from a stream of timestamped edges at the finest temporal granularity.
In particular, this work proposes the use of temporal walks as a basis to learn temporally valid node embeddings that capture the important temporal dependencies of the network at the finest most natural granularity (\eg, at a time scale of seconds or milliseconds).
This is in contrast to approximating the dynamic network as a sequence of static snapshot graphs $G_1,\ldots,G_t$ where each static snapshot graph represents all edges that occur between a user-specified discrete-time interval (\eg, day or week)~\cite{rossi2012dynamic-srl,soundarajan2016generating,sun2007graphscope}.
Besides the obvious loss of information, there are many other issues such as selecting an appropriate aggregation granularity which is known to be an important and challenging problem in itself that can lead to poor predictive performance or misleading results.
In addition, our approach naturally supports learning in \emph{graph streams} where edges arrive continuously over time (\eg, every second/millisecond)~\cite{aggarwal2011outlier,ahmed17streams,aggarwal2010dense,guha2012graph} and therefore can be used for a variety of applications requring real-time performance~\cite{pienta2015scalable,cai2012facilitating,ahmed2015interactive}.

We demonstrate the effectiveness of the proposed framework and generalized dynamic network embedding method for temporal link prediction in several real-world networks from a variety of application domains.
Overall, the proposed method achieves an average gain of $11.9\%$ across all methods and graphs. 
The results indicate that modeling temporal dependencies in graphs is important for learning appropriate and meaningful network representations.
In addition, any existing embedding method or deep graph model that use random walks can benefit from the proposed framework (\eg,~\cite{deepwalk,node2vec,struc2vec,ComE,ASNE,dong2017metapath2vec,ahmed17learning-attr-graphs,lee17-Deep-Graph-Attention}) as it serves as a basis for incorporating important temporal dependencies into existing methods.
Methods generalized by the framework are able to learn more meaningful and accurate time-dependent network embeddings that capture important properties from continuous-time dynamic networks.

Previous embedding methods and deep graph models that use random walks search over the space of random walks $\mathbb{S}$ on $G$,
whereas the class of models (continuous-time dynamic network embeddings) proposed in this work learn temporal embeddings by
searching over the space $\mathbb{S}_{T}$ of temporal random walks that obey time and thus $\mathbb{S}_{T}$ includes only \emph{temporally valid walks}.
See Figure~\ref{fig:space-of-random-walks} for intuition.
Informally, a \emph{temporal walk} $S_t$ from node $v_{i_{1}}$ to node $v_{i_{L+1}}$ is defined as a sequence of edges $\lbrace(v_{i_{1}}, v_{i_{2}}, t_{i_{1}})$, $(v_{i_{2}},v_{i_{3}}, t_{i_{2}}), \ldots, (v_{i_{L}},$ $v_{i_{L+1}}, t_{i_{L}})\rbrace$ such that $t_{i_{1}} \leq t_{i_{2}} \leq \ldots \leq t_{i_{L}}$.
A temporal walk represents a \emph{temporally valid} sequence of edges traversed in increasing order of edge times and therefore respect time.
For instance, suppose each edge represents a contact (\eg, email, phone call, proximity) between two entities, then a temporal random walk represents a feasible route for a piece of information through the dynamic network.
It is straightforward to see that existing methods that ignore time learn embeddings from a set of random walks that are not actually possible when time is respected and thus represent invalid sequences of events.
There is only a small overlap between $\mathbb{S}_T$ and $\mathbb{S}_D$ as shown in Figure~\ref{fig:space-of-random-walks} since only a small fraction of the space of walks in $\mathbb{S}_D$ are actually time-respecting (valid temporal walks).

\makeatletter
\global\let\tikz@ensure@dollar@catcode=\relax
\makeatother
\tikzstyle{every node}=[font=\large,line width=1.5pt]
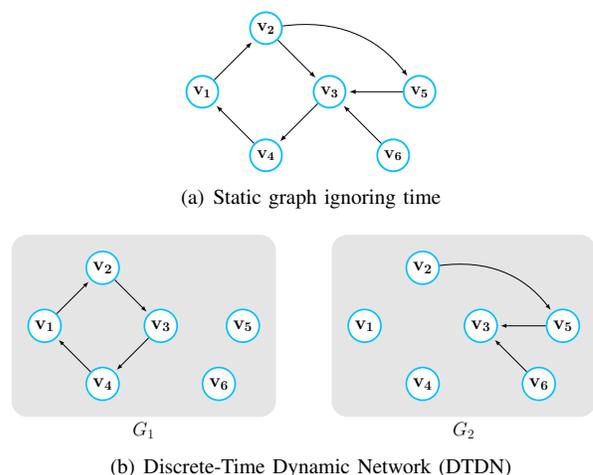
\begin{figure}[b!]
\vspace{-5mm}
\centering
\begin{center}

\subfigure[Static graph ignoring time]{
\label{fig:static-graph-example}
\scalebox{0.46}{
\centering
\begin{tikzpicture}[->,>=latex,shorten >=2.4pt,auto,node distance=2.6cm,thick,main node/.style={circle,draw=thelightblue,fill=white,draw,font=\sffamily\Large\bfseries}]

\node[main node] (3) {$\mathbf{v_2}$};
\node[main node] (1) [below left of=3] {$\mathbf{v_1}$};
\node[main node] (4) [below right of=1] {$\mathbf{v_4}$};
\node[main node] (2) [below right of=3] {$\mathbf{v_3}$};
\node[main node] (5) [right of=2] {$\mathbf{v_5}$};
\node[main node] (6) [below right of=2] {$\mathbf{v_6}$};

\tikzstyle{LabelStyle}=[below=3pt]
\path[every node/.style={font=\sffamily}] 

(1) edge [left] node [above left] {} (3) 
	(2)  edge [left] node[below right] {} (4)
	
(4) edge [left] node[below left] {} (1)

	(3) edge[bend left]  node[sloped,anchor=center,above] {} (5) 
	(5) edge  node[anchor=center,above] {} (2) 
	(6) edge  node[sloped,anchor=center,below] {} (2) 

	(3) edge [right] node[above right] {} (2);
\end{tikzpicture}
}
}

\tikzstyle{background-page}=[rectangle,
fill=gray!25,
inner sep=0.5cm,
rounded corners=5mm]     
\subfigure[Discrete-Time Dynamic Network (DTDN)]
{\label{fig:DTND-example} 
\begin{minipage}[t]{0.43\linewidth}
\scalebox{0.42}{
\centering
\begin{tikzpicture}[->,>=latex,shorten >=2.4pt,auto,node distance=2.6cm,thick,main node/.style={circle,draw=thelightblue,fill=white,draw,font=\sffamily\Large\LARGE\bfseries}]

\node[main node] (3) {$\mathbf{v_2}$};
\node[main node] (1) [below left of=3] {$\mathbf{v_1}$};
\node[main node] (4) [below right of=1] {$\mathbf{v_4}$};
\node[main node] (2) [below right of=3] {$\mathbf{v_3}$};
\node[main node] (5) [right of=2] {$\mathbf{v_5}$};
\node[main node] (6) [below right of=2] {$\mathbf{v_6}$};

\tikzstyle{LabelStyle}=[below=3pt]
\path[every node/.style={font=\sffamily}] 

(1) edge [left] node [above left] {} (3)

	(2)  edge [right] node[below right] {} (4)
	
(4) edge [left] node[below left] {} (1)

	(3) edge [right] node[above right] {} (2);

\begin{pgfonlayer}{background}
		\node [background-page, 
fit=(3) (1) (4) (2) (5) (6),
label=below:\fontsize{18}{20}\selectfont $G_1$ ] {};
\end{pgfonlayer}    

\end{tikzpicture}
}
\end{minipage}
\hspace{2mm}
\begin{minipage}[t]{0.43\linewidth}
\scalebox{0.42}{
\centering
\begin{tikzpicture}[->,>=latex,shorten >=2.4pt,auto,node distance=2.6cm,thick,main node/.style={circle,draw=thelightblue,fill=white,draw,font=\sffamily\Large\LARGE\bfseries}]

\node[main node] (3) {$\mathbf{v_2}$};
\node[main node] (1) [below left of=3] {$\mathbf{v_1}$};
\node[main node] (4) [below right of=1] {$\mathbf{v_4}$};
\node[main node] (2) [below right of=3] {$\mathbf{v_3}$};
\node[main node] (5) [right of=2] {$\mathbf{v_5}$};
\node[main node] (6) [below right of=2] {$\mathbf{v_6}$};

\tikzstyle{LabelStyle}=[below=3pt]
\path[every node/.style={font=\sffamily}] 

(3) edge[bend left]  node[sloped,anchor=center,above] {} (5) 
(5) edge  node[anchor=center,above] {} (2) 
(6) edge  node[sloped,anchor=center,below] {} (2); 

\begin{pgfonlayer}{background}
		\node [background-page, 
fit=(3) (1) (4) (2) (5) (6),
label=below:\fontsize{18}{20}\selectfont $G_2$] {};
\end{pgfonlayer}    

\end{tikzpicture}
}
\end{minipage}
}

\end{center}
\vspace{-4mm}
\caption{Representing the continuous-time dynamic network as a static graph or discrete-time dynamic network (DTDN). 
Noise and information loss occurs when the true dynamic network (Figure~\ref{fig:info-loss-example}) is approximated as a sequence of discrete static snapshot graphs $G_1,\ldots,G_t$ using a user-defined aggregation time-scale $s$ (temporal granularity).
Suppose the dynamic network in Figure~\ref{fig:info-loss-example} is used and $s=5$,
then $G_1$ includes all edges in the time-interval $[1,5]$ whereas $G_2$ includes all edges in $[6,10]$ and so on.
Notice that in the static snapshot graph $G_1$ the walk $v_4 \!\! \longrightarrow \! v_1 \!\! \longrightarrow \! v_2$ is still possible despite it being \emph{invalid} while the perfectly valid temporal walk $v_1 \!\! \longrightarrow \! v_2 \!\! \longrightarrow \! v_5$ is impossible.
Both cases are captured correctly without any loss using the notion of \emph{temporal walk} on the actual dynamic network.
}
\label{fig:info-discrete-time-model-loss-example}
\end{figure}

The sequence that links (events) occur in a network carries important information, \eg, if the event (link) represents an email communication sent from one user to another, then the state of the user who receives the email message changes in response to the email communication.
For instance, suppose we have two emails $e_i = (v_1,v_2)$ from $v_1$ to $v_2$ and $e_j=(v_2,v_3)$ from $v_2$ to $v_3$; and let $\mathcal{T}(v_1,v_2)$ be the time of an email $e_i = (v_1,v_2)$. 
If $\mathcal{T}(v_1,v_2) < \mathcal{T}(v_2,v_3)$ then the message $e_j = (v_2,v_3)$ may reflect the information received from the email communication $e_i=(v_1,v_2)$.
However, if $\mathcal{T}(v_1,v_2) > \mathcal{T}(v_2,v_3)$ then the message $e_j = (v_2,v_3)$ cannot contain any information communicated in the email $e_i=(v_1,v_2)$.
This is just one simple example illustrating the importance of modeling the actual sequence of events (email communications).
Embedding methods that ignore time are prone to many issues such as learning inappropriate node embeddings that do not accurately capture the dynamics in the network such as the real-world interactions or flow of information among nodes.
An example of information loss that occurs when time is ignored or the actual dynamic network is approximated using a sequence of discrete static snapshot graphs is shown in Figure~\ref{fig:info-loss-example} and~\ref{fig:info-discrete-time-model-loss-example}, respectively.
This is true for networks that involve the flow or diffusion of information through a network~\cite{lerman2010information,acemoglu2010spread,rossi2012dpr-dynamical}, networks modeling the spread of disease/infection~\cite{infect}, spread of influence in social networks (with applications to product adoption, viral marketing)~\cite{java2006modeling,domingos2005mining}, or more generally any type of dynamical system or diffusion process over a network~\cite{lerman2010information,acemoglu2010spread,rossi2012dpr-dynamical}.

The proposed approach naturally supports generating dynamic node embeddings for any pair of nodes at a specific time $t$. 
More specifically, given a newly arrived edge between node $i$ and $j$ at time $t$, we simply add the edge to the graph, perform a number of temporal random walks that contain those nodes, and then update the embedding vectors for those nodes (via a partial fast update) using only those walks.
In this case, there is obviously no need to recompute the embedding vectors for all such nodes in the graph as the update is very minor and an online partial update can be performed fast.
This includes the case where either node in the new edge has never been seen previously.
The above is a special case of our framework and is a trivial modification.
Notice that we can also obviously drop-out past edges as they may become stale.

\medskip\noindent\textbf{Summary of Main Contributions:}
This work makes three main contributions.
First, we described a new class of embeddings based on the notion of \emph{temporal walks}.
This notion can be used in a straightforward fashion to adapt other existing and/or future state-of-the-art methods for learning embeddings from temporal networks (graph streams).
Second, unlike previous work that learn embeddings using an approximation of the actual dynamic network (\ie, sequence of static graphs), we describe a new class of embeddings called \emph{continuous-time dynamic network embeddings} (CTDNE) that are learned directly from the graph stream.
CTDNEs avoid the issues and information loss that arise when time is ignored or the dynamic network (graph stream) is approximated as a sequence of discrete static snapshot graphs.
This new class of embeddings leverage the notion of
\emph{temporal walks} that captures the \emph{temporally valid interactions} (\eg, flow of information, spread of diseases) in the dynamic network (graph stream) in a lossless fashion.
As an aside, since these embeddings are learned directly from the graph stream at the finest granularity, they can also be learned in an online fashion, \ie, 
node embeddings are updated after every new edge (or batch of edges).
Finally, we describe a framework for learning them based on the notion of \emph{temporal walks}.
The proposed framework provides a basis for generalizing existing (or future state-of-the-art) embedding methods that use the traditional notion of random walks over static or discrete approximation of the actual dynamic network.

\newcommand{\subsec}[1]{\medskip\noindent\textbf{#1:}\;}

\section{Related work} \label{sec:related-work}
\noindent

\subsec{Representation Learning in Static Networks}
The node embedding problem has received considerable attention from the research community in recent years.\footnote{In the time between our shorter CTDNE paper from early 2018~\cite{CTDNE} and this papers original submission, there have been a number of closely related follow-up works. 
For temporal clarity, these works are not reviewed or compared against in detail.}
See~\cite{rossi12jair} for an early survey on representation learning in relational/graph data.
The goal is to learn encodings (embeddings, representations, features) that capture key properties about each node such as their 
role in the graph based on their structural characteristics (\ie, roles capture distinct structural properties, \eg, hub nodes, bridge nodes, near-cliques)~\cite{rossi2014roles}
or community (\ie, communities represent groups of nodes that are close together in the graph based on proximity, cohesive/tightly connected nodes)~\cite{ng2002spectral,pons2006computing}.
Since nodes that share similar roles (based on structural properties) or communities (based on proximity, cohesiveness) 
are grouped close to each other in the embedding space, one can easily use the learned embeddings for tasks such as ranking~\cite{page1998pagerank}, community detection~\cite{ng2002spectral,pons2006computing}, role embeddings~\cite{rossi2014roles,ahmed2017edgeroles}, link prediction~\cite{liu2010link}, and node classification~\cite{rossi2012dynamic-srl}.

Many of the techniques that were initially proposed for solving the node embedding problem were based on graph factorization~\cite{ahmedWWW13,Belkin02laplacianeigenmaps,grarep}. 
More recently, the skip-gram model~\cite{skipgram-old} was introduced in the natural language processing domain to learn vector representations for words. Inspired by skip-gram's success in language modeling, various methods~\cite{deepwalk,node2vec,line} have been proposed to learn node embeddings using skip-gram by treating a graph as a ``document." 
Two of the more notable methods, DeepWalk~\cite{deepwalk} and node2vec~\cite{node2vec}, use random walks to sample an ordered sequence of nodes from a graph. The skip-gram model can then be applied to these sequences to learn node embeddings. 

\subsec{Representation Learning in Dynamic Networks}
Researchers have also tackled the problem of node embedding in more complex graphs, including attributed networks~\cite{ASNE}, heterogeneous networks~\cite{dong2017metapath2vec} and dynamic networks~\cite{rossi2013dbmm-wsdm,zhou2018dynamic,li2017attributed}.
However, the majority of the work in the area still fail to consider graphs that evolve over time (\ie temporal graphs). 
A few work have begun to explore the problem of learning node embeddings from temporal networks~\cite{rossi2013dbmm-wsdm,hisano2016semi,kamra2017dgdmn, zhu2016scalable,saha2018models,rahman2018dylink2vec}. 
All of these approaches \emph{approximate} the dynamic network as a sequence of discrete static snapshot graphs, which are fundamentally different from the class of continuous-time dynamic network embedding methods introduced in this work.
Notably, this work is the first to propose \emph{temporal random walks} for embeddings as well as \emph{CTDN embeddings} that use temporal walks to capture the actual temporally valid sequences observed in the CTDN; and thus avoids the issues and information loss that arises when embedding methods simply ignore time or use discrete static snapshot graphs (See Figure~\ref{fig:info-discrete-time-model-loss-example} for one example).
Furthermore, we introduce a unifying framework that can serve as a basis for generalizing other random walk based deep learning (\eg,~\cite{lee17-Deep-Graph-Attention}) and embedding methods (\eg,~\cite{struc2vec,node2vec,ComE,ASNE,dong2017metapath2vec,hamilton2017inductive})
for learning more appropriate time-dependent embeddings from temporal networks. 
In contrast, previous work has simply introduced new approaches for temporal networks~\cite{hisano2016semi} and therefore they focus on an entirely different problem than the one in this work which is a general framework that can be leveraged by other non-temporal approaches.

Temporal graph smoothing of a sequence discrete static snapshot graphs was proposed for classification in dynamic networks~\cite{rossi2012dynamic-srl}. 
The same approach has also been used for deriving role-based embeddings from dynamic networks~\cite{rossi2012role-www,rossi2013dbmm-wsdm}.
More recently, these techniques have been used to derive more meaningful embeddings from a sequence of discrete static snapshot graphs~\cite{bonner2018temporal,singer2019node,saha2018models,rahman2018dylink2vec}.
All of these approaches model the dynamic network as a sequence of discrete static snapshot graphs, which is fundamentally different from the class of continuous-time dynamic network embedding methods introduced in this work.
Table~\ref{table:qual-comp} provides a qualitative comparison of CTDNE methods to existing static methods or DTDNE methods that approximate the dynamic network as a discrete sequence of static snapshot graphs.

\begin{table}[t!]
\centering
\renewcommand{\arraystretch}{1.10}
\caption{Comparison of Different Classes of Embedding Methods}
\label{table:qual-comp}
\vspace{-2.5mm}
\footnotesize
\setlength{\tabcolsep}{2.9pt} 
\begin{tabularx}{1.0\linewidth}{l@{} cc ccc cHH @{}}
\multicolumn{8}{@{}p{1.0\linewidth}}{
\scriptsize
Comparison of CTDNE methods to existing methods categorized as either static methods (that ignore time) or DTDNE methods that approximate the actual dynamic network using a sequence of discrete static snapshot graphs.
Does the method:
use the actual dynamic network at the finest temporal granularity, \eg, seconds or ms (or do they use discrete static approximations of the dynamic network);
temporally valid;
use temporal bias/smoothing functions to give more importance to recent or temporally recurring information;
and does it naturally support graph streams and the streaming/online setting in general where data is continuously arriving over time and embeddings can be incrementally updated in an online fashion.
}\\
\toprule
& {\footnotesize \bf Temporally } & & {\footnotesize \bf Finest} 		& {\footnotesize \bf Temporal} &&&  \\ 
& {\footnotesize \bf valid} 			   & & {\footnotesize \bf granularity} & {\footnotesize \bf bias/smoothing} && {\footnotesize \bf Streaming} && 
\\ 
\midrule
\textsf{Static} & \xmark  & & \xmark & \xmark && \xmark & \\
\textsf{DTDNE} & \xmark  & & \xmark & \cmark && \xmark & \\
\textsf{CTDNE} & \cmark & & \cmark & \cmark  && \cmark & \\  
\bottomrule
\end{tabularx}
\vspace{-2mm}
\end{table}

\subsec{Temporal Networks}
More recently, there has been significant research in developing network analysis and machine learning methods for modeling temporal networks.
Temporal networks have been the focus of recent research including 
node classification in temporal networks~\cite{rossi2012dynamic-srl},
temporal link prediction~\cite{dunlavy2011temporal}, 
dynamic community detection~\cite{cazabet2014dynamic}, 
dynamic mixed-membership role models~\cite{fu2009dynamic,rossi2012role-www,rossi2013dbmm-wsdm}, 
anomaly detection in dynamic networks~\cite{ranshous2015dynamic-net-anomaly-survey},
influence modeling and online advertisement~\cite{goyal2010learning}, 
finding important entities in dynamic networks~\cite{rossi2012dpr-dynamical,OMadadhain2005}, and
temporal network centrality and measures~\cite{holme2012temporal,beres2018temporal}.

\subsec{Random Walks}
Random walks on graphs have been studied for decades~\cite{lovasz1993random}.
The theory underlying random walks and their connection to eigenvalues and other fundamental properties of graphs are well-understood~\cite{chung2007random}.
Our work is also related to uniform and non-uniform random walks on graphs~\cite{lovasz1993random,chung2007random}.
Random walks are at the heart of many important applications such as ranking~\cite{page1998pagerank}, community detection~\cite{ng2002spectral,pons2006computing}, 
recommendation~\cite{bogers2010movie}, 
link prediction~\cite{liu2010link}, and
influence modeling~\cite{java2006modeling}.
search engines~\cite{lassez:latentlinks}, 
image segmentation~\cite{grady2006random}, routing in wireless sensor networks~\cite{servetto2002constrained}, and time-series forecasting~\cite{rossi2012dpr-dynamical}.
These applications and techniques may also benefit from the proposed class of embeddings that are based on the notion of \emph{temporal random walks}.
Recently, Ahmed~\etal~\cite{ahmed17attrRandomWalks} proposed the notion of \emph{attributed random walks} that can be used to generalize existing methods for inductive learning and/or graph-based transfer learning tasks.
In future work, we will investigate combining both attributed random walks and temporal random walks~\cite{tremblay2001temporal} to derive even more powerful embeddings.

\section{Continuous-Time Dynamic Embeddings}
\label{sec:streaming-network-embeddings}
\noindent
While previous work uses discrete approximations of the dynamic network (\ie, a sequence of discrete static snapshot graphs), this paper proposes an entirely new direction that instead focuses on learning embeddings directly from the graph stream using only temporally valid information.

In this work, instead of approximating the dynamic network as a sequence of discrete static snapshot graphs defined as $G_1, \ldots, G_T$ where $G_i=(V, E_t)$ and $E_t$ are the edges active between the timespan $[t_{i-1},t_i]$, we model the \emph{temporal interactions} in a lossless fashion as a \emph{continuous-time dynamic network} (CTDN) defined formally as:
\begin{Definition}[\sc Continuous-Time Dynamic Network] \label{eq:cont-time-dynamic-network}
Given a graph $G=(V,E_T,\mathcal{T})$, let $V$ be a set of vertices, and $E_T \subseteq V \times V \times \RR^{+}$ be a set of temporal edges between vertices in $V$, and $\mathcal{T} : E \rightarrow \RR^{+}$ is a function that maps each edge to a corresponding timestamp.
At the finest granularity, each edge $e_i = (u,v,t) \in E_T$ may be assigned a unique time $t \in \RR^{+}$.
\end{Definition}\noindent
In continuous-time dynamic networks (\ie, temporal networks, graph streams)~\cite{holme2012temporal}, edges occur over a time span $\mathcal{T} \subseteq \mathbb{T}$ where $\mathbb{T}$ is the temporal domain.\footnote{The terms temporal network, graph stream, and continuous-time dynamic network are used interchangeably.}
For continuous-time systems $\mathbb{T}=\RR^{+}$.
In such networks, a \emph{valid} walk is defined as a sequence of nodes connected by edges with non-decreasing timestamps~\cite{nicosia2013graph}. 
In other words, if each edge captures the time of contact between two entities, then a (valid temporal) walk may represent a feasible route for a piece of information. 
More formally, 
\begin{Definition}[\sc Temporal Walk]\label{def:temporal-walk}
A temporal walk from $v_1$ to $v_k$ in $G$ is a sequence of vertices $\langle v_1, v_2, \cdots, v_k \rangle$ such that $\langle v_i, v_{i+1} \rangle \in E_T$ for $1 \leq i < k$, and $\mathcal{T}(v_i, v_{i+1}) \leq \mathcal{T}(v_{i+1}, v_{i+2})$ for $1 \leq i < (k-1)$. 
For two arbitrary vertices $u$, $v \in V$, we say that $u$ is \textit{temporally connected} to $v$ if there exists a temporal walk from $u$ to $v$.
\end{Definition}
\noindent
The definition of temporal walk echoes the standard definition of a walk in static graphs but with an additional constraint that requires the walk to respect time, that is,
edges must be traversed in increasing order of edge times.
As such, temporal walks are naturally asymmetric~\cite{xuan2003computing,ferreira2007evaluation,tremblay2001temporal}.
Modeling the dynamic network in a continuous fashion makes it completely trivial to add or remove edges and nodes.
For instance, suppose we have a new edge $(v,u,t)$ at time $t$, then we can sample a small number of temporal walks ending in $(v,u)$ and perform a fast partial update to obtain the updated embeddings (See Section~\ref{sec:time-preserving-embeddings} for more details).
This is another advantage to our approach compared to previous work that use discrete static snapshot graphs to approximate the dynamic network.
Note that performing a temporal walk forward through time is equivalent to one backward through time. 
However, for the streaming case (online learning of the embeddings) where we receive an edge $(v,u,t)$ at time $t$, then we sample a temporal walk backward through time. 
A \emph{temporally invalid walk} is a walk that does not respect time.
Any method that uses temporally invalid walks or approximates the dynamic network as a sequence of static snapshot graphs is said to have \emph{temporal loss}.

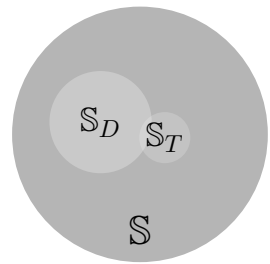
\begin{figure}[h!]
\vspace{0mm}
\centering

\begin{center}
\scalebox{0.85}{
\begin{tikzpicture}
\begin{scope}[blend group = soft light]
\fill[gray!70]   ( 90:1.5) circle (2);
\fill[gray!50] (110:1.8) circle (0.8);
\fill[gray!70] (75:1.5) circle (0.4);
\end{scope}
\node [font=\Large] {\fontsize{18}{20}\selectfont $\mathbb{S}$};
\node at ( 110:1.8)   {\fontsize{15}{17}\selectfont $\mathbb{S}_D$};
\node at ( 75:1.5)   {\fontsize{15}{17}\selectfont $\mathbb{S}_T$};

\end{tikzpicture}
}
\end{center}

\vspace{-3mm}
\caption{
Space of all possible random walks $\mathbb{S}$ (up to a fixed length $L$) including 
(i) the space of temporal (time-obeying) random walks denoted as $\mathbb{S}_T$ that capture the temporally valid flow of information (or disease, etc.) in the network without any loss
and (ii) the space of random walks that are possible when the dynamic network is approximated as a sequence of discrete static snapshot graphs denoted as $\mathbb{S}_{D}$.
Notably, there is a very small overlap between $\mathbb{S}_T$ and $\mathbb{S}_D$ since only a small fraction of the walks in $\mathbb{S}_D$ are actually time-respecting (valid temporal walks).
}
\label{fig:space-of-random-walks}
\vspace{0mm}
\end{figure}
We define a new type of embedding for dynamic networks (graph streams) called continuous-time dynamic network embedding (CTDNEs).
\begin{Definition}[\sc Continuous-Time Dynamic Network Embedding]\label{def:ctdne-problem}
Given a dynamic network $G=(V,E_T,\mathcal{T})$ (graph stream), the goal is to learn a function $f : V \rightarrow \RR^{D}$ that maps nodes in the continuous-time dynamic network (graph stream) $G$ to $D$-dimensional time-dependent embeddings
using only data that is temporally valid (\eg, temporal walks defined in Definition~\ref{def:temporal-walk}).
\end{Definition}\noindent
Unlike previous work that ignores time or \emph{approximates} the dynamic network as a sequence of discrete static snapshot graphs $G_1, \ldots, G_t$, CTDNEs proposed in this work are learned from temporal random walks  that capture the true temporal interactions (\eg, flow of information, spread of diseases, etc.) in the dynamic network in a lossless fashion.
CTDNEs (or simply dynamic node embeddings) can be learned incrementally or in a streaming fashion where embeddings are updated in real-time as new edges arrive.
For this new class of dynamic node embeddings, we describe a general framework for learning such temporally valid embeddings from the graph stream in Section~\ref{sec:framework}.

\section{Framework}
\label{sec:framework}
\noindent
While Section~\ref{sec:streaming-network-embeddings} formally introduced the new class of embeddings investigated in this work, this section describes a general framework for deriving them based on the notion of $\text{\emph{temporal walks}}$.
The framework has two main interchangeable components that can be used to \emph{temporally bias} the learning of the dynamic node embeddings.
We describe each component in Section~\ref{sec:selection-of-start-time} and~\ref{sec:temporal-random-walk}. 
In particular, the CTDNE framework generates \emph{(un)biased temporal random walks} from CTDNs 
that are then used in Section~\ref{sec:time-preserving-embeddings} for deriving time-dependent embeddings that are learned from temporally valid node sequences that capture in a lossless fashion the actual flow of information or spread of disease in a network.
It is straightforward to use the CTDNE framework for temporal networks where edges are active only for a specified time-period.

\begin{figure}[h!]
\vspace{-2mm}
\centering
\hspace{-4mm}
\includegraphics[width=0.6\linewidth]{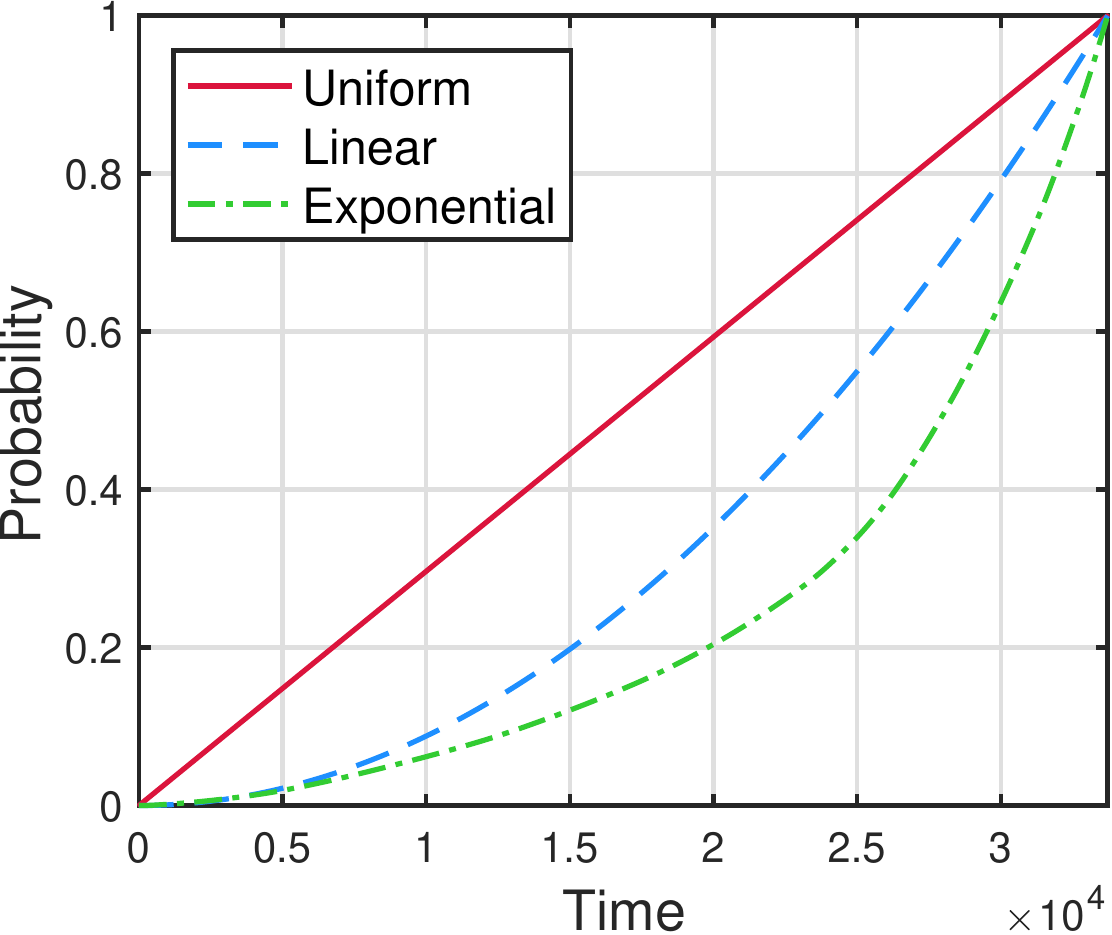}
\caption{
Example initial edge selection cumulative probability distributions (CPDs) for each of the variants investigated (uniform, linear, and exponential).
Observe that exponential biases the selection of the initial edge towards those occurring more recently than in the past, whereas linear lies between exponential and uniform.
}
\label{fig-initial-edge-selection-fb-forum}
\end{figure}

\subsection{Initial Temporal Edge Selection} 
\label{sec:selection-of-start-time}
This section describes approaches to temporally bias the temporal random walks by selecting the initial temporal edge to begin the temporal random walk.
In general, each temporal walk starts from a temporal edge $e_i \in E_T$ at time $t=\mathcal{T}$ selected from a distribution $\mathbb{F}_s$. 
The distribution used to select the initial temporal edge can either be uniform in which case there is no bias 
or the selection can be temporally biased using an arbitrary weighted (non-uniform) distribution for $\mathbb{F}_s$.
For instance, to learn node embeddings for the temporal link prediction task, we may want to begin more temporal walks from edges closer to the current time point as the events/relationships in the distant past may be less predictive or indicative of the state of the system now.
Selecting the initial temporal edge in an unbiased fashion is discussed in Section~\ref{sec:selection-of-start-time-unbiased} whereas strategies that temporally bias the selection of the initial edge are discussed in Section~\ref{sec:selection-of-start-time-biased}.
In the case of learning CTDNEs in an online fashion, we do not need to select the initial edge since we simply sample a number of temporal walks that end at the new edge. 
See Section~\ref{sec:time-preserving-embeddings} for more details on learning CTDNEs in an online fashion.

\subsubsection{Unbiased} \label{sec:selection-of-start-time-unbiased} 
In the case of initial edge selection, each edge $e_i=(v,u,t) \in E_T$ has the same probability of being selected:
	\begin{equation}\label{eq:uniform-edge}
	\Pr(e) = 1 / |E_T| 
	\end{equation}\noindent
This corresponds to selecting the initial temporal edge using a uniform distribution.

\subsubsection{Biased} \label{sec:selection-of-start-time-biased} 
We describe two techniques to temporally bias the selection of the initial edge that determines the start of the temporal random walk.
In particular, we select the initial temporal edge using a temporally weighted distribution based on exponential and linear functions.
However, the proposed continuous-time dynamic network embedding framework is flexible with many interchangeable components and therefore can easily support other temporally weighted distributions for selecting the initial temporal edge.

\medskip\noindent\textbf{Exponential:}
We can also bias initial edge selection using an exponential distribution, in which case each edge $e \in E_T$ is assigned the probability:
\begin{equation}\label{eq:exponential-dist}
	\Pr(e) = \frac{\exp\big[ \mathcal{T}(e)-t_{\min}]}{\sum_{e^\prime \in E_T} \, \exp\big[ \mathcal{T}(e^\prime)-t_{\min}]}
\end{equation}\noindent
where $t_{\min}$ is the minimum time associated with an edge in the dynamic graph. 
This defines a distribution that heavily favors edges appearing later in time.

\medskip\noindent\textbf{Linear:}  
When the time difference between two time-wise consecutive edges is large, it can sometimes be helpful to map the edges to discrete time steps. Let $\eta : E_T \rightarrow \mathbb{Z}^{+}$ be a function that sorts (in ascending order by time) the edges in the graph. In other words $\eta$ maps each edge to an index with $\eta(e) = 1$ for the earliest edge $e$. In this case, each edge $e \in \eta(E_T)$ will be assigned the probability:

\begin{equation}\label{eq:linear-dist}
\Pr(e) = \frac{\eta(e)}{\sum_{e^\prime \in E_T} \eta(e^\prime)}
\end{equation}\noindent

\textcolor{red}{
See Figure~\ref{fig-initial-edge-selection-fb-forum} for examples of the uniform, linear, and exponential variants.
}

\subsection{Temporal Random Walks} 
\label{sec:temporal-random-walk}
\noindent
After selecting the initial edge $e_i = (u, v, t)$ at time $t$ to begin the temporal random walk (Section~\ref{sec:selection-of-start-time}) using $\mathbb{F}_s$, 
how can we perform a temporal random walk starting from that edge?
We define the set of temporal neighbors of a node $v$ at time $t$ as follows:
\begin{Definition}[\sc Temporal Neighborhood]\label{def:temporal-neighbor}
	The set of temporal neighbors of a node $v$ at time $t$ denoted as $\Gamma_t(v)$ are:
	\begin{equation}\label{eq:potential-neighbors-at-time-t}
	\Gamma_t(v) = \big\{(w, t^\prime) \,\, | \,\, e=(v,w, t^\prime) \in E_T \, \wedge \mathcal{T}(e) > t \big\} 
	\end{equation}
\end{Definition}
\noindent
Observe that the same neighbor $w$ can appear multiple times in $\Gamma_t(v)$ since multiple temporal edges can exist between the same pair of nodes. 
See Figure~\ref{fig:temporal-neighbors} for an example.
The next node in a temporal random walk can then be chosen from the set $\Gamma_t(v)$.
Here we use a second distribution $\mathbb{F}_\Gamma$ to \emph{temporally bias} the neighbor selection. 
Again, this distribution can either be uniform, in which case no bias is applied, or more intuitively biased to consider time. 
For instance, we may want to bias the sampling strategy towards walks that exhibit smaller ``in-between" time for consecutive edges. That is, for each consecutive pair of edges $(u, v, t)$, and $(v, w, t+k)$ in the random walk, we want $k$ to be small. 
For temporal link prediction on a dynamic social network, restricting the ``in-between" time allows us to sample walks that do not group friends from different time periods together. 
As an example, if $k$ is small we are likely to sample the random walk sequence $(v_1, v_2, t), (v_2, v_3, t+k)$ which makes sense as $v_1$ and $v_3$ are more likely to know each other since $v_2$ has interacted with them both recently. On the other hand, if $k$ is large we are unlikely to sample the sequence. This helps to separate people that $v_2$ interacted with during very different time periods (\textit{e.g.} high-school and graduate school) as they are less likely to know each other. 

\makeatletter
\global\let\tikz@ensure@dollar@catcode=\relax
\makeatother
\tikzstyle{every node}=[font=\large,line width=1.5pt]
\begin{figure}[h!]
\centering
\begin{center}

\subfigure[Neighborhood $\Gamma(v_2)$]
{\label{fig:neighborhood-example}
\scalebox{0.55}{
\centering
\begin{tikzpicture}[->,>=latex,shorten >=2.4pt,auto,node distance=2.6cm,thick,
main node/.style={circle,draw=thelightblue,fill=white,draw,font=\sffamily\Large\bfseries},
inactive node/.style={circle,draw=gray!150,fill=white,draw,font=\sffamily\Large\bfseries,text=gray!150}]

\node[main node] (2) {$\mathbf{v_3}$};
\node[main node] (1) [below left of=2] {$\mathbf{v_2}$};
\node[main node] (3) [left of=1] {$\mathbf{v_1}$};
\node[main node] (4) [below right of=1] {$\mathbf{v_5}$};
\node[main node] (5) [right of=1] {$\mathbf{v_4}$};

\node[main node] (6) [above left of=1] {$\mathbf{v_8}$};
\node[main node] (8) [below left of=1] {$\mathbf{v_6}$};

\tikzstyle{LabelStyle}=[below=3pt]
\path[every node/.style={font=\sffamily}] 

	(3) edge [thick,line width=0.6mm,left] node [above left] {\textbf{t=6}} (1) 
	
	(1)  edge [right] node[above right] {} (6)
(1)  edge [right] node[above left] {} (8)

	(1)  edge [right] node[above right] {} (5)
	
(1) edge [left] node[below left] {} (4)

	(1) edge [right] node[above left] {} (2);
\end{tikzpicture}
}
}
\hspace{4mm}
\subfigure[Temporal neigh. $\Gamma_{t}(v_2)$]
{\label{fig:temporal-neighborhood-example}
\scalebox{0.55}{
\centering
\begin{tikzpicture}[->,>=latex,shorten >=2.4pt,auto,node distance=2.6cm,thick,
main node/.style={circle,draw=thelightblue,fill=white,draw,font=\sffamily\Large\bfseries},
inactive node/.style={circle,draw=gray!150,fill=white,draw,font=\sffamily\Large\bfseries,text=gray!150}]

\node[main node] (2) {$\mathbf{v_3}$};
\node[main node] (1) [below left of=2] {$\mathbf{v_2}$};
\node[main node] (3) [left of=1] {$\mathbf{v_1}$};
\node[main node] (4) [below right of=1] {$\mathbf{v_5}$};
\node[main node] (5) [right of=1] {$\mathbf{v_4}$};

\node[inactive node] (6) [above left of=1] {$\mathbf{v_8}$};
\node[inactive node] (8) [below left of=1] {$\mathbf{v_6}$};

\tikzstyle{LabelStyle}=[below=3pt]
\path[every node/.style={font=\sffamily}] 

	(3) edge [thick,line width=0.6mm,left] node [above left] {\textbf{t=6}} (1) 
	
	(1)  edge [draw=gray!150,text=black, dashed,right] node[above right] {4} (6)
(1)  edge [draw=gray!150,text=black, dashed,right] node[above left] {1} (8)

	(1)  edge [right] node[above right] {7} (5)
	
(1) edge [left] node[below left] {9} (4)

	(1) edge [right] node[above left] {8,10} (2);
\end{tikzpicture}
}
}
\end{center}

\vspace{-4mm}
\caption{
Temporal neighborhood of a node $v_2$ at time $t=6$ denoted as $\Gamma_t(v_2)$.
Notice that $\Gamma_t(v_2) = \{v_4, v_3, v_5, v_3\}$ is an ordered multiset where the temporal neighbors are sorted in ascending order by time with the nodes more recent appearing first. 
Moreover, the same node can appear multiple times (\eg, a user sends another user multiple emails, or an association/event occurs multiple times between the same entities).
This is in contrast to the definition of neighborhood used by previous work that is not parameterized by time, \eg, $\Gamma(v_2) = \{v_3, v_4, v_5, v_6, v_8\}$ or $\Gamma(v_2) = \{v_3, v_3, v_4, v_5, v_6, v_8\}$ if multigraphs are supported.
}
\label{fig:temporal-neighbors}
\vspace{-2mm}
\end{figure}
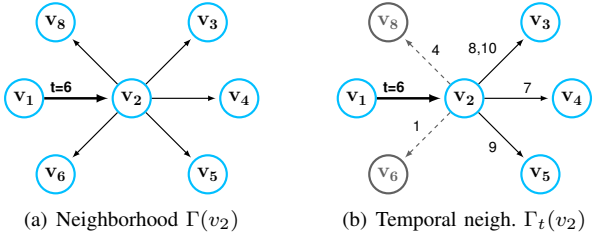

\subsubsection{Unbiased} \label{sec:temporal-random-walk-unbiased} 
For unbiased temporal neighbor selection, given an arbitrary edge $e = (u, v, t)$, each temporal neighbor $w \in \Gamma_t(v)$  of node $v$ at time $t$ has the following probability of being selected:
\begin{equation}\label{eq:uniform-neighbor}
\Pr(w) = 1 / |\Gamma_t(v)|
\end{equation}\noindent

\subsubsection{Biased} \label{sec:temporal-random-walk-biased} 
We describe two techniques to bias the temporal random walks by sampling the next node in a temporal walk via temporally weighted distributions based on exponential and linear functions.
However, the continuous-time dynamic network embedding framework is flexible and can easily be used with other application or domain-dependent \emph{temporal bias functions}.

\medskip\noindent\textbf{Exponential:} 
When exponential decay is used, we formulate the probability as follows. Given an arbitrary edge $e = (u, v, t)$, each temporal neighbor $w \in \Gamma_t(v)$ has the following probability of being selected:
\begin{equation}\label{eq:exponential-penalty}
\Pr(w) = \frac{\exp\!\big[ \tau(w) - \tau(v)\big]}{\sum_{w^\prime \in \Gamma_t(v)} \exp\!\big[ \tau(w^\prime) - \tau(v) \big]}
\end{equation}\noindent
Note that we abuse the notation slightly here and use $\tau$ to mean the mapping to the corresponding time. 
This is similar to the exponentially decaying probability of consecutive contacts observed in the spread of computer viruses and worms~\cite{holme2012temporal}.

\medskip\noindent\textbf{Linear:}  
Here, we define $\delta : V \times \RR^{+}  \rightarrow \mathbb{Z}^{+}$ as a function which sorts temporal neighbors in descending order time-wise. The probability of each temporal neighbor $w \in \Gamma_t(v)$ of node $v$ at time $t$ is then defined as:
\begin{equation}\label{eq:linear-penalty}
\Pr(w) = \frac{\delta(w)}{\sum_{w^\prime \in \Gamma_t(v)} \delta(w^\prime)}
\end{equation}\noindent 
This distribution biases the selection towards edges that are closer in time to the current node.

\subsubsection{Temporal Context Windows}
Since temporal walks preserve time, it is possible for a walk to run out of \emph{temporally valid} edges to traverse.
Therefore, we do not impose a strict length on the temporal random walks.
Instead, we simply require each temporal walk to have a minimum length $\omega$ (in this work, $\omega$ is equivalent to the context window size for skip-gram \cite{skipgram-old}). 
A maximum length $L$ can be provided to accommodate longer walks. 
A temporal walk $\mathcal{S}_{t_i}$ with length $|\mathcal{S}_{t_i}|$ is considered valid iff 
\[
\omega \leq |\mathcal{S}_{t_i}| \leq L
\]
Given a set of temporal random walks $\{ \mathcal{S}_{t_1}, \mathcal{S}_{t_2}, \cdots, \mathcal{S}_{t_k}\}$, we define the temporal context window count $\beta$ as the total number of context windows of size $\omega$ that can be derived from the set of temporal random walks. 
Formally, this can be written as:
\begin{equation} \label{eq:stopping-criterion}
\beta \, = \sum_{i=1}^{k} \big( |\mathcal{S}_{t_i}| - \omega + 1\big)
\end{equation}
\noindent
When deriving a set of temporal walks, we typically set $\beta$ to be a multiple of $N = |V|$.
Note that this is only an implementation detail and is not important for Online CTDNEs.

\begin{figure*}[t!]
\centering
\includegraphics[width=0.28\linewidth]{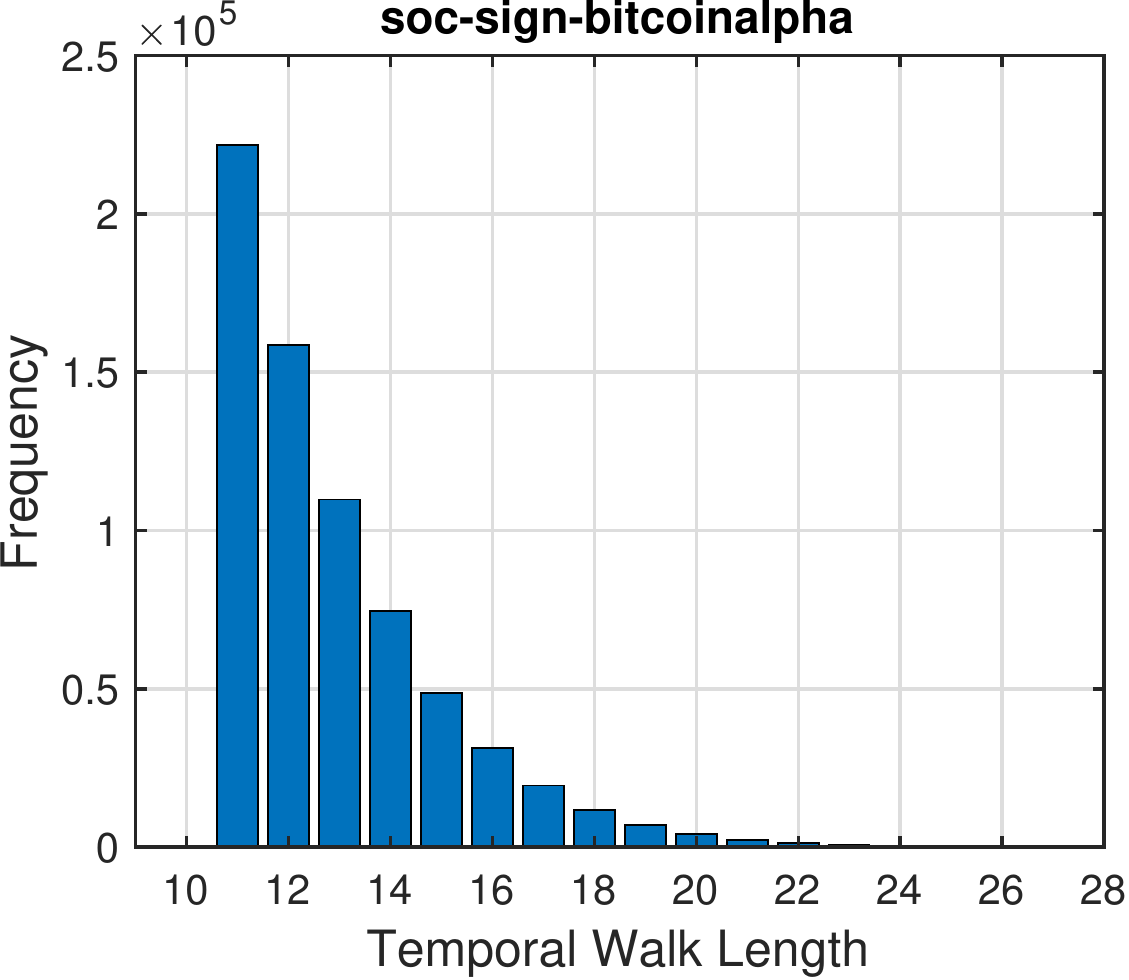}
\hspace{4mm}
\includegraphics[width=0.28\linewidth]{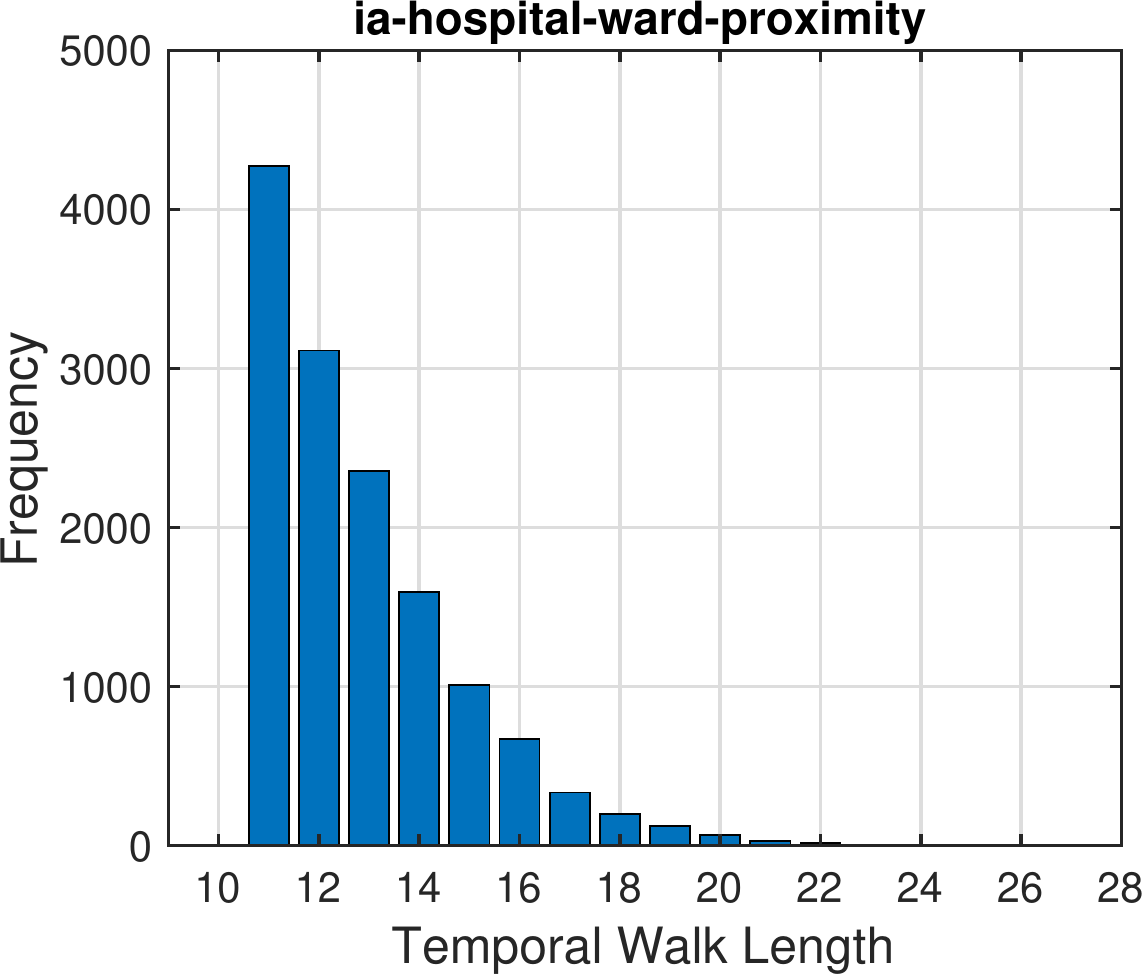}
\hspace{4mm}
\includegraphics[width=0.28\linewidth]{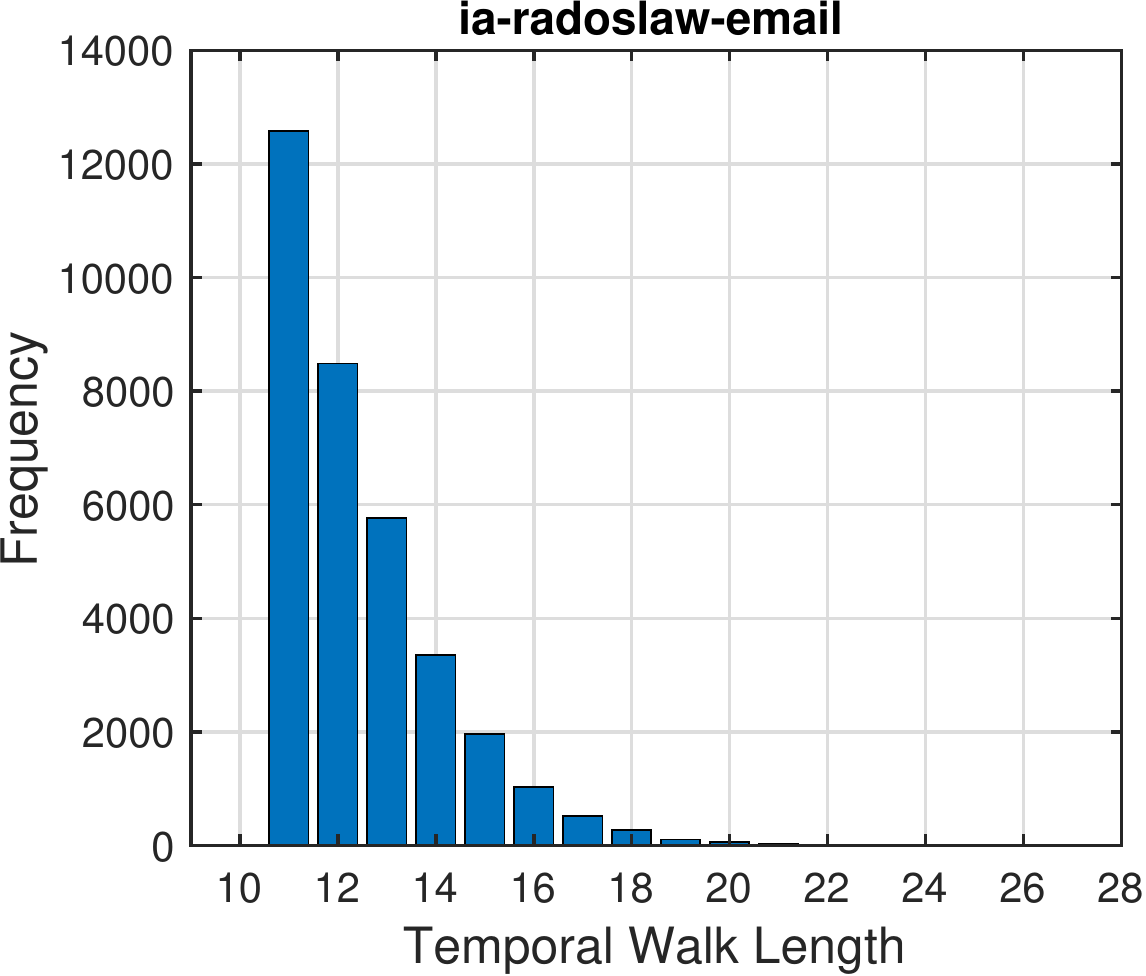}
\vspace{-1mm}
\caption{Frequency of \emph{temporal random walks} by length}
\label{fig:temporal-walk-length-freq}
\end{figure*}

\subsection{Learning Dynamic Node Embeddings}
\label{sec:time-preserving-embeddings}
\noindent
Given a temporal walk $\mathcal{S}_{t}$, we can now formulate the task of learning time-preserving node embeddings in a CTDN as the optimization problem:
\begin{align} \label{eq:obj-func}
\max_{f} \; \log \Pr \big(\,W_T = \{v_{i-\omega},\cdots,v_{i+\omega} \} \setminus v_i \;|\; f(v_i) \big)
\end{align}\noindent
where 
$f : V \rightarrow \RR^{D}$ is the node embedding function, 
$\omega$ is the context window size for optimization, and
\[
W_T = \{v_{i-\omega},\cdots,v_{i+\omega} \}
\]\noindent
such that 
\[
\mathcal{T}(v_{i-\omega},v_{i-\omega+1}) < \cdots < \mathcal{T}(v_{i+\omega-1},v_{i+\omega})
\]\noindent 
is an arbitrary temporal context window $W_{T} \subseteq S_t$.
For tractability, we assume conditional independence between the nodes of a temporal context window when observed with respect to the source node $v_i$. That is:
\begin{align} \label{eq:conditional-indep}
\Pr \big(\,W_T | f(v_i) \big) = \prod_{v_{i+k} \in W_T} \Pr \big(v_{i+k} | f(v_i) \big)
\end{align} \noindent
We can model the conditional likelihood of every source-nearby node pair $(v_i, v_j)$ as a softmax unit parameterized by a dot product of their feature vectors:
\begin{align}\label{eq:cond-ll}
\Pr \big(\,v_j | f(v_i) \big) = \frac{\exp\!\big[ f(v_j) \cdot f(v_i)\big]}{\sum_{v_k \in V} \exp\!\big[ f(v_k) \cdot  f(v_i) \big]}
\end{align}\noindent
Using Eq.~\ref{eq:conditional-indep}-~\ref{eq:cond-ll}, the optimization problem in Eq.~\ref{eq:obj-func} reduces to:
\begin{align}\label{eq:obj-func-simplifies}
\max_{f} \; \sum_{v_i \in V} \Bigg( - \log Z_i + \sum_{v_{j} \in W_T}  f(v_j) \cdot f(v_i) \Bigg)
\end{align}\noindent
where the term $Z_i = \sum_{v_j \in V} \exp\!\big[ f(v_i) \cdot f(v_j) \big]$ can be approximated by negative sampling.
Given a graph $G$, let $\mathbb{S}$ be the space of all possible random walks on $G$ and let $\mathbb{S}_{T}$ be the space of all temporal random walks on $G$.
It is straightforward to see that the space of temporal random walks $\mathbb{S}_{T}$ is contained within $\mathbb{S}$, and $\mathbb{S}_{T}$ represents only a tiny fraction of possible random walks in $\mathbb{S}$.
Existing methods sample a set of random walks $\mathcal{S}$ from $\mathbb{S}$ whereas this work focuses on sampling a set of \emph{temporal random walks} $\mathcal{S}_t$ from $\mathbb{S}_{T} \subseteq \mathbb{S}$ (Fig.~\ref{fig:space-of-random-walks}).
In general, the probability of an existing method sampling a temporal random walk from $\mathbb{S}$ by chance is extremely small and therefore the vast majority of random walks sampled by these methods represent sequences of events between nodes that are invalid (not possible) when time is respected.
\smallskip
\begin{Claim}
Fix $L>0$, then 
$|\mathbb{S}| \gg |\mathbb{S}_D| \gg |\mathbb{S}_T|$.
\end{Claim}
\smallskip
\noindent
Therefore, previous methods that learn embeddings from random walks are unlikely to generate \emph{temporally valid sequences} of events/interactions between nodes that are actually possible when time is respected.

{\algrenewcommand{\alglinenumber}[1]{\fontsize{6.5}{7}\selectfont#1 }
	\newcommand{\multiline}[1]{\State \parbox[t]{\dimexpr\linewidth-\algorithmicindent}{#1\strut}}
	\begin{figure}[h!]
		\vspace{-2mm}
		\centering
		\begin{algorithm}[H]
			\caption{\,\small Continuous-Time Dynamic Network Embeddings
}
			\label{alg:temporal-node2vec}
			{
				\begin{spacing}{1.15}
					\fontsize{7.5}{8.5}\selectfont
					\begin{algorithmic}[1]
						\vspace{-1.3mm}
						\Require 
						a dynamic network (graph stream) 
						$G = (V,\E_T,\mathcal{T})$,
						temporal context window count $\beta$, 
						context window size $\omega$,
						embedding dimensions $D$
						\smallskip
						 
						\State Initialize number of temporal context windows $C = 0$ 
						\While {$\beta - C > 0$ }
						\State Sample an edge $e_{t}\!=\!(v,u)$ via $\mathbb{F}_s$ (or use new edge at time $t$)
						\State $t \leftarrow \mathcal{T}(e_{t})$ 
						
						\State $S_t = \textsc{TemporalWalk}(G, e_{t}, t, L, \omega + \beta - C - 1)$ 
						\label{algline:obtain-temporal-walk}
						\If {$|S_t| > \omega$}
						\State Add the \emph{temporal walk} $S_t$ to $\mathcal{S}_T$ \label{algline:add-temporal-walk-to-set}
						\State $C \leftarrow C + (|S_t| - \omega + 1)$
						\EndIf
						
						\EndWhile
						\State $\mZ = \textsc{StochasticGradientDescent}(\omega, D, \mathcal{S}_T)$ \label{algline:SGD-with-temporal-walks} \Comment{update embeddings} 
						\State \textbf{return}  
						\emph{dynamic} node embeddings $\mZ$ \label{algline:return-learned-representation-matrix} 
					\end{algorithmic}
			\end{spacing}}
		\end{algorithm}
		\vspace{-2mm}
\end{figure}}

{\algrenewcommand{\alglinenumber}[1]{\fontsize{6.5}{7}\selectfont#1 }
	\newcommand{\multiline}[1]{\State \parbox[t]{\dimexpr\linewidth-\algorithmicindent}{#1\strut}}
	\begin{figure}[h!]
		\vspace{-9.2mm}
		\begin{algorithm}[H]
			\caption{\,\small 
				Temporal Random Walk
			}
			\label{alg:temporal-random-walk}{
				\begin{spacing}{1.15}
					\fontsize{7.5}{8.5}\selectfont
					\begin{algorithmic}[1]
						\vspace{-1.3mm}
						\Procedure{TemporalWalk}{$G^{\prime}$, 
						$e=(s,r)$,
							$t$, 
							$C$}
						
						\State Set $i \leftarrow s$ and initialize temporal walk $S_t = \big[\, s, r \,\big]$  \label{algline:temporal-walk-init-walk-and-add-start-node-function}
						\For{$p = 1$ {\bf to} $C - 1$} \label{algline:temporal-walk-for}
						
						\State $\Gamma_t(i) = \big\{(w, t^\prime) \,\, | \,\, e=(i,w, t^\prime) \in E_T \, \wedge \mathcal{T}(i) > t \big\} $ \label{algline:temporal-walk-get-neighbors}
						
						\If {$|\Gamma_t(i)| > 0$}
						
						\State Select node $j$ from distribution $\mathbb{F}_\Gamma (\Gamma_t(i))$ \label{algline:temporal-walk-alias-sample}
						\State Append $j$ to $S_t$ \label{algline:temporal-walk-add-node-function-to-list}
						\State Set $t \leftarrow \mathcal{T}(i,j)$ and set $i \leftarrow j$

						\Else \; terminate temporal walk
						\EndIf
						\EndFor \label{algline:temporal-walk-for-end}
				
						\State \textbf{return} temporal walk $S_t$ of length $|S_t|$ rooted at node $s$ \label{algline:temporal-walk-return-temporal-walk}
						\EndProcedure
					\end{algorithmic}
			\end{spacing}}
		\end{algorithm}
		\vspace{-2mm}
	\end{figure}
}

We summarize the procedure to learn time-preserving embeddings for CTDNs in Algorithm~\ref{alg:temporal-node2vec}.
Our procedure in Algorithm~\ref{alg:temporal-node2vec} generalizes the Skip-Gram architecture to learn continuous-time dynamic network embeddings (CTDNEs). 
However, the framework can easily be used for other deep graph models that leverage random walks (\eg,~\cite{lee17-Deep-Graph-Attention}) 
as the temporal walks can serve as input vectors for neural networks.
There are many methods that can be adapted to learn CTDN embeddings using \emph{temporal random walks} (\eg, node2vec~\cite{node2vec}, struc2vec~\cite{struc2vec}, role2vec~\cite{role2vec}) and the proposed framework is not tied to any particular approach.

We point out that Algorithm~\ref{alg:temporal-node2vec} is useful for prediction tasks where the goal is to learn a model using all data up to time $t$ for prediction of a future discrete or real-valued attribute or state (\eg, if a link exists or not). 
Since this work evaluates CTDNEs for link prediction, we include it mainly for the reader to understand one evaluation strategy using CTDNE.
However, other applications may require online incremental learning and updating of the embeddings in a streaming fashion as new edges arrive.
Recall that CTDNE naturally supports such streaming settings where edges (or new nodes) arrive continuously over time~\cite{ahmed17streams} and the goal is to update the embeddings in real-time via fast efficient updates.

In Algorithm~\ref{alg:CTDNE-online}, we present an online CTDNE learning framework for incrementally updating the node embeddings as new edges arrive over time from the edge stream.
Consider an edge stream $e_1, e_2, \ldots, e_k,\ldots, e_{t-1}, e_{t}, \ldots$ with timestamped edges.
Suppose a new edge $(v,u,t)$ arrives at time $t$ from the edge stream (Line~\ref{algline:online-CTDNE-while-edge-arrives}).
Then we immediately update the graph by adding the edge $(v,u,t)$ to $E \leftarrow E \cup \{(v,u,t)\}$ as shown in Line~\ref{algline:online-CTDNE-add-edge-and-nodes-if-needed}.\footnote{At this point, we can also remove any stale edges as well, \eg, edges that occurred in the distant past defined by some $\Delta t$.}
If either $v$ or $u$ are new nodes, \ie, $v \not\in V$ or $u \not\in V$, then we simply set $V \leftarrow V \cup \{v,u\}$.
Notice that if $v,u \in V$ then $V \leftarrow V \cup \{v,u\}$ in Line~\ref{algline:online-CTDNE-add-edge-and-nodes-if-needed} has no impact. 
The next step is to sample a set of temporal walks $\mathcal{S}_{t}$ with the constraint that each temporal walk ends at the new edge $(v,u,t)$ from the edge stream (Line~\ref{algline:online-CTDNE-sample-temporal-walks}).
We obtain temporal walks that end in $(v,u,t)$ by reversing the temporal walk and going backwards through time as shown in Figure~\ref{fig:online-temporal-walk}.
This enables us to easily obtain a set of temporal walks that include the new edge, which will be used for incrementally updating the embeddings.
Furthermore, since the goal is to obtain temporal walks that include the new edge, then we know $(v,u,t)$ will be at the end of the temporal walk (since by definition no other edge could have appeared after it), and we simply obtain the temporal walk by going backwards through time. 
Finally, we incrementally update the appropriate node embeddings using only the sampled temporal walks $\mathcal{S}_{t}$ ending at $(v,u,t)$ at time $t$ (Line~\ref{algline:online-CTDNE-update-embeddings}).
In this work, we use online SGD updates (online word2vec)~\cite{kaji2017incremental,peng2017incrementally,luo2015online,li2017psdvec} to incrementally learn the embeddings as new edges arrive.
However, other incremental optimization schemes can easily be used as well~\cite{duchi2011adaptive,flaxman2005online,zhao2012fast,schraudolph2007stochastic,ge2015escaping,ying2008online}.
While Algorithm~\ref{alg:CTDNE-online} assumes the graph stream is infinite, the current and most recently updated embeddings $\vz_1, \vz_2, \ldots, \vz_N$ can be obtained at any time $t$.

Concept drift is naturally handled by the framework since we incrementally update embeddings upon the arrival of each edge in the stream using walks that are temporally valid.
Hence, the context and resulting embedding of a node changes temporally as the graph evolves over time.
Furthermore, we can relax the requirement of updating the embeddings after every new edge, and instead, we can wait until a fixed number of edges arrive before updating the embeddings or wait until a fixed amount of time elapses.
We call such an approach batched CTDNE updating.
The only difference in Algorithm~\ref{alg:CTDNE-online} is that instead of performing an update immediately, we would wait until one of the above conditions become true and then perform a batch update.
We can also drop edges that occur in the distant past or that have a very small weight.

{\algrenewcommand{\alglinenumber}[1]{\fontsize{6.5}{7}\selectfont#1 }
	\newcommand{\multiline}[1]{\State \parbox[t]{\dimexpr\linewidth-\algorithmicindent}{#1\strut}}
	\begin{figure}[t!]
		\vspace{-2mm}
		\centering
		
		\begin{algorithm}[H]
			\caption{\,
			Online Learning of Node Embeddings from Edge Streams (Online CTDNE)
			}
			\label{alg:CTDNE-online}
			{
				\begin{spacing}{1.15}
					\fontsize{8.0}{9.0}\selectfont
					\begin{algorithmic}[1]
						\vspace{-0.5mm}
						\Require 
						a dynamic network (graph stream)
						$G$, 
						embedding dimensions $D$
						\Ensure dynamic node embeddings $\mZ$ at time $t$
						\smallskip

						\While{new edge $(v,u,t)$ arrives at time $t$ from edge stream} \label{algline:online-CTDNE-while-edge-arrives}
							\State Add edge $(v,u,t)$ to $E \leftarrow E \cup \{(v,u,t)\}$ and $V \leftarrow V \cup \{v,u\}$ 
							\label{algline:online-CTDNE-add-edge-and-nodes-if-needed}

							\State Sample temporal walks 
							$\mathcal{S}_{t}$ 
							ending in edge $(v,u,t)$ \label{algline:online-CTDNE-sample-temporal-walks}
							
							\State Update embeddings via online SGD/word2vec using only $\mathcal{S}_{t}$	\label{algline:online-CTDNE-update-embeddings}
							
						\EndWhile
						\vspace{0.2mm}
					\end{algorithmic}
			\end{spacing}}
		\end{algorithm}
		\vspace{-7mm}
\end{figure}
}

\vspace{-4mm}
\subsection{Hyperparameters}
\noindent
While other methods have a lot of hyperparameters that require tuning such as node2vec~\cite{node2vec}, the proposed framework has a single hyperparameter that requires tuning.
Note that since the framework is general and flexible with many interchangeable components, there is of course the possibility of introducing additional hyperparameters depending on the approaches used to bias the temporal walks.

\begin{figure}[t!]
\centering
\includegraphics[width=0.46\linewidth]{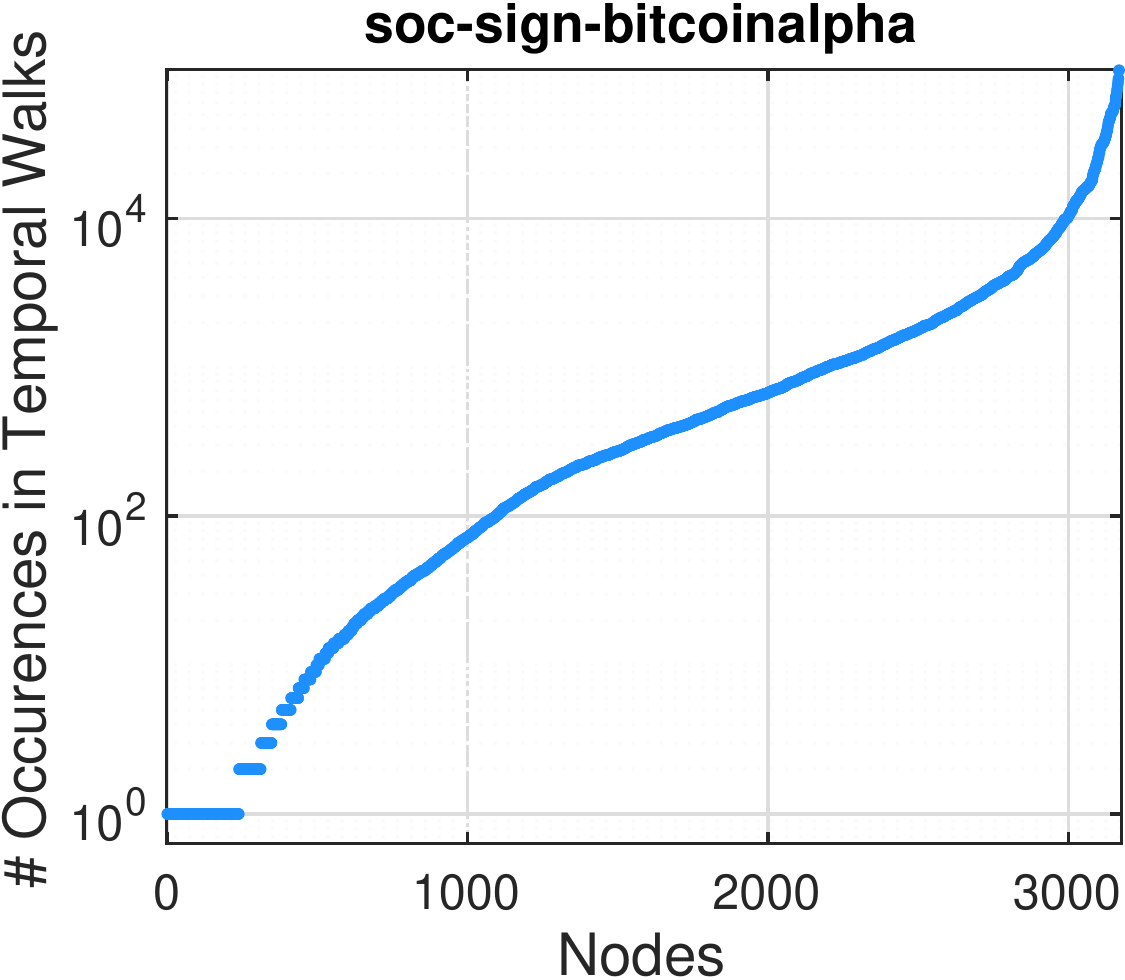}
\hfill
\includegraphics[width=0.46\linewidth]{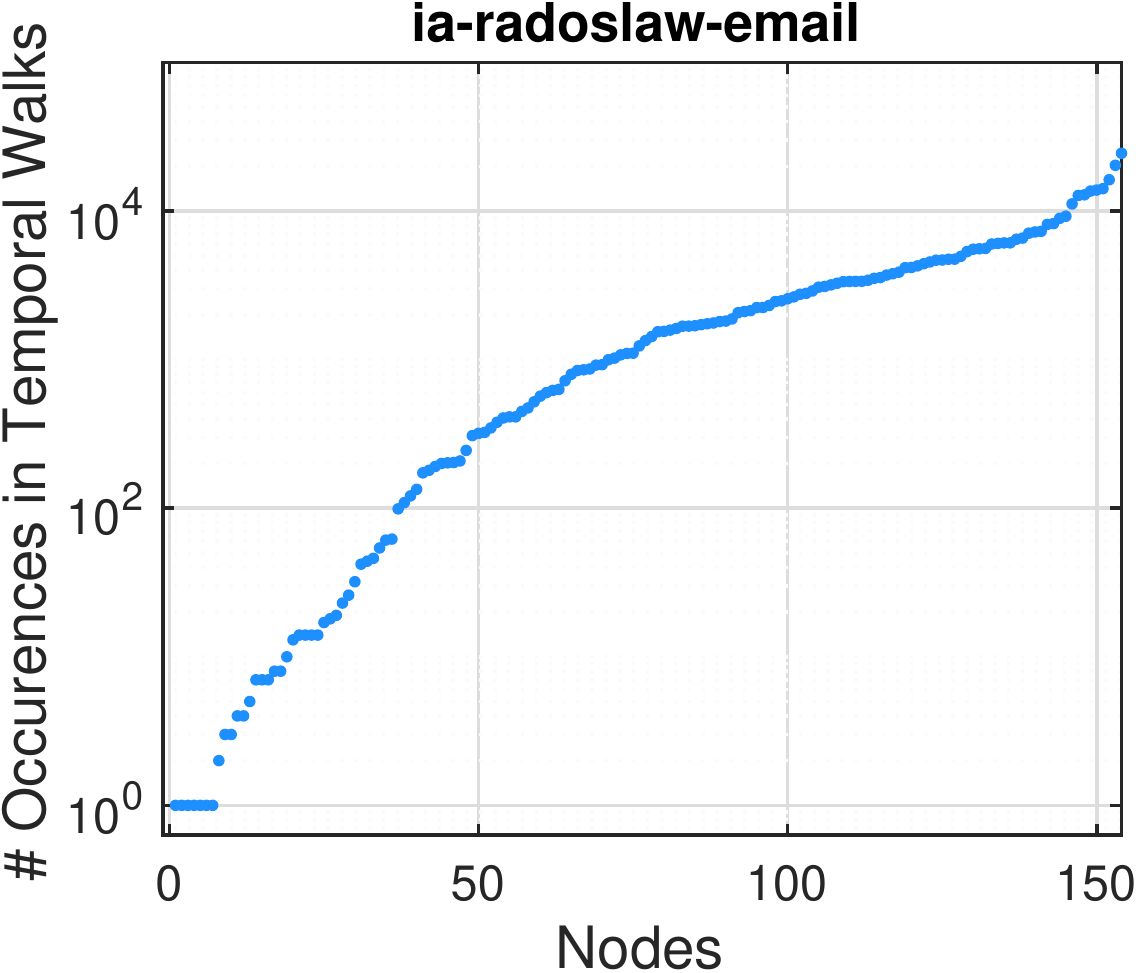}
\vspace{-1mm}
\caption{Number of occurrences of each node in the set of sampled temporal walks.}
\label{fig:node-occur-temporal-walks}
\end{figure}

\begin{figure}[t!]
\centering
\includegraphics[width=0.46\linewidth]{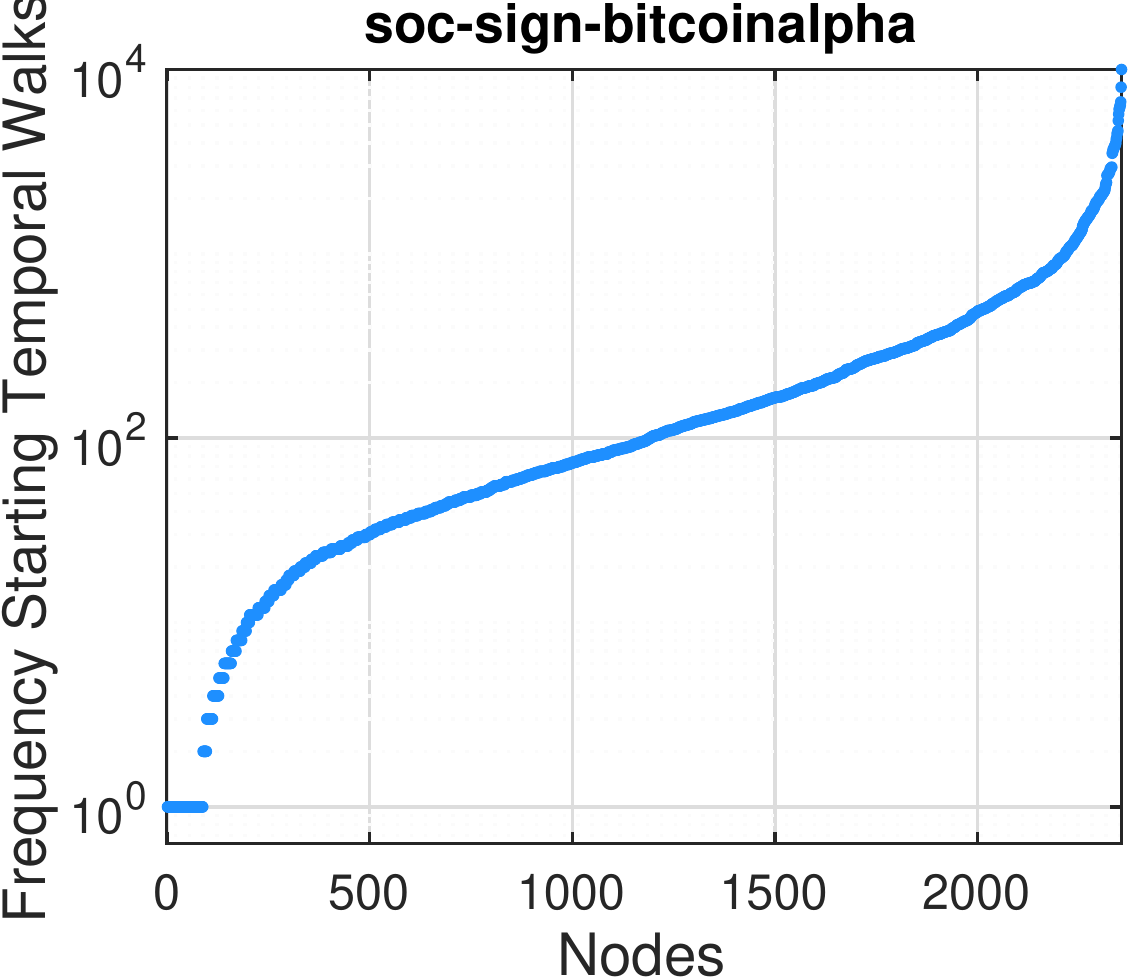}
\hfill
\includegraphics[width=0.46\linewidth]{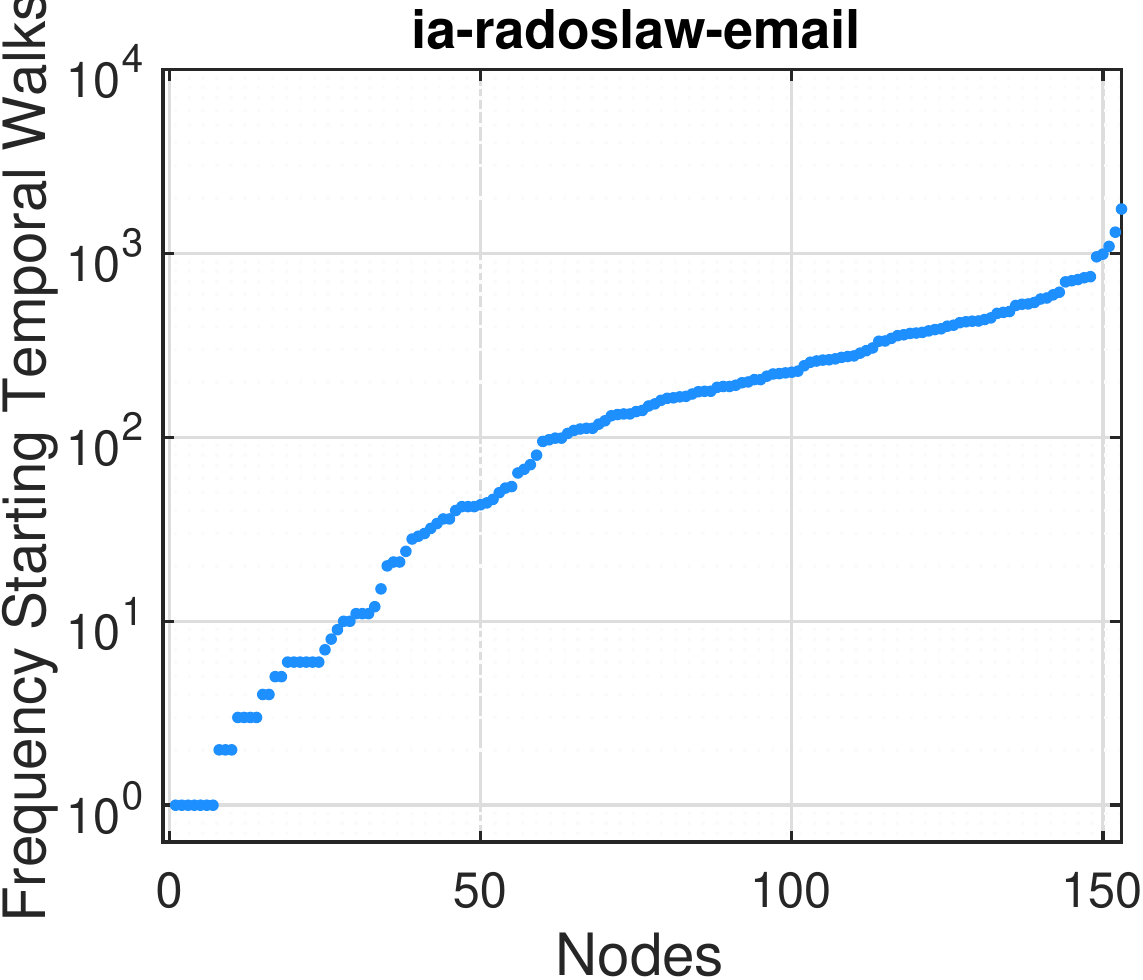}
\vspace{-1mm}
\caption{Frequency of starting a temporal random walk from each node. 
Unlike previous approaches that sample a fixed number of random walks for each node, the proposed framework samples an edge between two nodes to obtain a timestamp to begin the temporal random walk.
}
\label{fig:node-starting-temporal-walk-freq}
\end{figure}

\medskip\noindent\textbf{Arbitrary temporal walk length}: 
Unlike walks in static graphs, temporal walks in the proposed framework can be of any arbitrary length.
In particular, the user does not need to select the length of the walks to sample as required by static embedding methods~\cite{node2vec,deepwalk}, among the many other hyperparameters required by such methods.
As an aside, the temporal context size $\omega$ is not specific to the framework, but arises from the base embedding method that we used.
For instance, suppose node2vec/deepwalk is used as the base embedding method in the proposed framework, then $\omega$ is simply the context/window size, and therefore, the only requirement on the length of the walk is that it is at least as large as $\omega$, which ensures at least one temporal context can be generated from it.
This is obviously better than node2vec/deepwalk, which requires selecting at least $L$, $R$, and $\omega$. 
Figure~\ref{fig:node-occur-temporal-walks} investigates the number of times each node appears in the sampled temporal walks.
We also study the frequency of starting a temporal random walk from each node in Figure~\ref{fig:node-starting-temporal-walk-freq}.

\section{Theoretical Analysis} \label{sec:complexity}
\noindent
Let $N=|V|$ denote the number of nodes, 
$M=|E_T|$ be the number of edges, 
$D = $ dimensionality of the embedding,
$R = $ the number of temporal walks per node, 
$L = $ the maximum length of a temporal random walk, 
and $\Delta = $ the maximum degree of a node. 
Recall that while $R$ is not required, we use it here since the number of temporal random walks $|\mathcal{S}_T|$ is a multiple of the number of nodes $N=|V|$ and thus can be written as $RN$ similar to previous work.

\subsection{Time Complexity}
\noindent
\begin{Lemma}
The time complexity for learning CTDNEs using the generalized Skip-gram architecture in Section~\ref{sec:time-preserving-embeddings} is
\begin{equation}\label{eq:time-complexity-CTDNE-biased}
\mathcal{O}(M + N (R \log M + R{L}\Delta + D))
\end{equation}\noindent
and the time complexity for learning CTDNEs with \emph{unbiased} temporal random walks (uniform) is:
\begin{equation}\label{eq:time-complexity-DTDNE-biased}
\mathcal{O}(N (R \log M + R{L}\log \Delta + D))
\end{equation}\noindent
\end{Lemma}
\noindent
\noindent\textsc{Proof}. 
The time complexity of each of the three steps is provided below.
We assume the exponential variant is used for both $\mathbb{F}_s$ and $\mathbb{F}_{\Gamma}$ since this CTDNE variant is the most computationally expensive among the nine CTDNE variants expressed from using uniform, linear, or exponential for $\mathbb{F}_s$ and $\mathbb{F}_{\Gamma}$.
Edges are assumed to be ordered by time such that $\mathcal{T}(e_1) \leq \mathcal{T}(e_2) \leq \cdots \leq \mathcal{T}(e_{M})$.
Similarly, the neighbors of each node are also ordered by time.

\textbf{Initial Temporal Edge Selection:}
To derive $\mathbb{F}_s$ for any of the variants used in this work (uniform, linear, exponential) it takes $\mathcal{O}(M)$ time since each variant can be computed with a single or at most two passes over the edges.
Selecting an initial edge via $\mathbb{F}_s$ takes $\mathcal{O}(\log M)$ time.
Now $\mathbb{F}_s$ is used to select the initial edge for each temporal random walk $S_{t} \in \mathcal{S}_T$ and thus an initial edge is selected $RN=|\mathcal{S}_T|$ times.
This gives a total time complexity of $\mathcal{O}(M + RN \log M)$.\footnote{Note for uniform initial edge selection, the time complexity is linear in the number of temporal random walks $\mathcal{O}(RN)$.}

\textbf{Temporal Random Walks:} 
After the initial edge is selected, the next step is to select the next temporally valid neighbor from the set of temporal neighbors $\Gamma_{t}(v)$ of a given node $v$ at time $t$ using a (weighted) distribution $\mathbb{F}_{\Gamma}$ (\eg, uniform, linear, exponential). 
Note $\mathbb{F}_{\Gamma}$ must be computed and maintained for each node.
Given a node $v$ and a time $t_{*}$ associated with the previous edge traversal in the temporal random walk, 
the first step in any variant (uniform, linear, exponential; Section~\ref{sec:temporal-random-walk}) is to obtain the ordered set of temporal neighbors $\Gamma_{t}(v) \subseteq \Gamma(v)$ of node $v$ that occur after $t_{*}$.
Since the set of all temporal neighbors is already stored and ordered by time, we only need to find the index of the neighbor $w \in \Gamma(v)$ with time $t>t_{*}$ as this gives us $\Gamma_{t}(v)$.
Therefore, $\Gamma_{t}(v)$ is derived in $\log |\Gamma(v)|$ via a binary search over the ordered set $\Gamma(v)$.
In the worst case, $\mathcal{O}(\log \Delta)$ where $\Delta = \max_{v \in V} |\Gamma(v)|$ is the maximum degree.
After obtaining $\Gamma_{t}(v) \subseteq \Gamma(v)$, we derive $\mathbb{F}_{\Gamma}$ in $\mathcal{O}(\Delta)$ time when $d_v = \Delta$.
Now, selecting the next temporally valid neighbor according to $\mathbb{F}_{\Gamma}$ takes $\mathcal{O}(\log \Delta)$ for exponential and linear and $o(1)$ for uniform.
For the uniform variant, we select the next temporally valid neighbor in $o(1)$ constant time by $j \sim \textrm{UniformDiscrete}\{1,2,\ldots,|\Gamma_t(v)|\}$ and then obtain the selected temporal neighbor by directly indexing into $\Gamma_t(v)$.
Therefore, the time complexity to select the next node in a biased temporal random walk is
$\mathcal{O}(\log \Delta + \Delta) = \mathcal{O}(\Delta)$ in the worst case and $\mathcal{O}(\log \Delta)$ for unbiased (uniform).

For a temporal random walk of length ${L}$, the time complexity is $\mathcal{O}({L}\Delta)$ for a biased walk with linear/exponential and $\mathcal{O}({L} \log \Delta)$ for an unbiased walk. 
Therefore, the time complexity for $RN$ biased temporal random walks of length ${L}$ is $\mathcal{O}(RN{L}\Delta)$ in the worst case and $\mathcal{O}(RN{L}\log \Delta)$ for unbiased.

\textbf{Learning Time-dependent Embeddings:} 
For the Skip-Gram-based generalization given in Section~\ref{sec:time-preserving-embeddings}, the time complexity per iteration of Stochastic Gradient Descent (SGD) is $\mathcal{O}(ND)$ where $D \ll N$.
While the time complexity of a single iteration of SGD is less than a single iteration of Alternating Least Squares (ALS)~\cite{pilaszy2010fast}, SGD requires more iterations to obtain a good enough model and is sensitive to the choice of learning rate~\cite{yun2014nomad,oh2015fast}.
Moreover, SGD is more challenging to parallelize compared to ALS~\cite{pilaszy2010fast} or Cyclic Coordinate Descent (CCD)~
\cite{kim2014algorithms,rossi2015dsaa-pcmf}.
Nevertheless, the choice of optimization scheme depends on the objective function of the embedding method generalized via the CTDNE framework.

\subsection{Space Complexity}
\noindent
Storing the $F_{s}$ distribution takes $\mathcal{O}(M)$ space.
The temporal neighborhoods do not require any additional space (as we simply store an index).
Storing $\mathbb{F}_{\Gamma}$ takes $\mathcal{O}(\Delta)$ (which can be reused for each node in the temporal random walk).
The embedding matrix $\mZ$ takes $\mathcal{O}(ND)$ space.
Therefore, the space complexity of CTDNEs is $\mathcal{O}(M + ND + \Delta) = \mathcal{O}(M + ND)$.
This obviously holds in the online stream setting where edges arrive continuously over time and updates are made in an online fashion since this is a special case of the more general CTDNE setting.

\begin{table}[b!]
\vspace{-3mm}
\centering
\renewcommand{\arraystretch}{1.15}
\fontsize{8}{9}\selectfont
\setlength{\tabcolsep}{6.0pt}
\caption{Dynamic Network Data and Statistics.}
\vspace{-2.5mm}
\label{table:dynamic-network-stats}
\begin{tabular}{r l ll c H@{}}
\multicolumn{6}{@{}p{0.94\linewidth}}{\footnotesize 
Let $|E_T|$ = number of \emph{temporal edges};
$\bar{d}$ = average temporal node degree; 
and $d_{\max}$ = max temporal node degree.
} \\
\toprule
& & & & \textbf{Timespan}   \\
\textbf{Dynamic Network} &  $|E_T|$ & $\bar{d}$ & $d_{\max}$ & \textbf{(days)} 
\\
\midrule
\text{ia-contact} &  28.2K & 206.2 & 2092 &  3.97  \\
\text{ia-hypertext} & 20.8K & 368.5 & 1483 &  2.46  \\
\text{ia-enron-employees} & 50.5K & 669.8 & 5177 & 1137.55  \\
\text{ia-radoslaw-email} & 82.9K & 993.1 & 9053 & 271.19  \\
\text{ia-email-EU} & 332.3K & 674.1 & 10571 & 803.93  \\
\text{ fb-forum} & 33.7K & 75.0 & 1841 & 164.49  \\
\text{soc-bitcoinA} & 24.1K & 12.8 & 888 & 1901.00  \\
\text{soc-wiki-elec} & 107K & 30.1 & 1346 & 1378.34  \\
\bottomrule
\end{tabular}
\end{table}

\section{Experiments} 
\label{sec:exp}
\noindent
The experiments are designed to investigate the effectiveness of the proposed \emph{continuous-time dynamic network embeddings} (CTDNE) framework for prediction.
To ensure the results and findings of this work are significant and meaningful,
we investigate a wide range of temporal networks from a variety of application domains with fundamentally different structural and temporal characteristics.
A summary of the dynamic networks used for evaluation and their statistics are provided in Table~\ref{table:dynamic-network-stats}. 
All networks investigated are continuous-time dynamic networks with $\mathbb{T} = \RR^{+}$. 
For these dynamic networks, the time scale of the edges is at the level of seconds or milliseconds, \ie, the edge timestamps record the time an edge occurred at the level of seconds or milliseconds (finest granularity given as input). 
Our approach uses the finest time scale available in the graph data as input.
All data is from NetworkRepository~\cite{nr} and is easily accessible for reproducibility.

We designed the experiments to answer four important questions.
First, are \emph{continuous-time dynamic network embeddings} (CTDNEs)
more useful than embeddings from methods that ignore time?
Second, how do the different embedding methods from the CTDNE framework compare?
Third, are CTDNEs better than embeddings learned from a sequence of discrete snapshot graphs that approximate the edge stream (DTNE methods)?
Finally, can we incrementally learn node embeddings fast using the online CTDNE framework?

\subsection{Experimental setup}
\noindent
Since this work is the first to learn embeddings over an edge stream (CTDN), there are no methods that are directly comparable.
Nevertheless, we first compare CTDNE against node2vec~\cite{node2vec}, DeepWalk~\cite{deepwalk}, and LINE~\cite{line}.
For node2vec, we use the same hyperparameters ($D=128$, $R=10$, $L=80$, $\omega = 10$) 
and grid search over $p,q\in \{0.25, 0.50, 1, 2, 4\}$ as mentioned in~\cite{node2vec}. 
The same hyperparameters are used for DeepWalk (with the exception of $p$ and $q$).
Unless otherwise mentioned, CTDNE methods use $\omega = 10$ and $D=128$.
For LINE, we also use $D=128$ with 2nd-order-proximity and number of samples $T=$ 60 million.

\begin{table}[h!]
\centering
\small
\fontsize{8}{9}\selectfont
\renewcommand{\arraystretch}{1.15}
\setlength{\tabcolsep}{2.0pt}
\caption{Results for Temporal Link Prediction (AUC).}
\label{table:link-pred-results}
\vspace{-2.4mm}
\begin{tabularx}{1.00\linewidth}{r cc X c r}
		\toprule
		\textbf{Dynamic Network} & \textbf{DeepWalk}
		& \textbf{Node2Vec}
		& \textbf{LINE}
		& \textbf{CTDNE} 
		& (\textsc{Gain})
		\\
		\midrule
		
		\text{ia-contact} & \text{0.845} & \textrm{0.874} & \textrm{0.736} & \textbf{0.913}  &  (\text{+10.37\%}) \\
		\text{ia-hypertext} & \text{0.620} & \textrm{0.641} & \textrm{0.621} & \textbf{0.671} &  (\text{+6.51\%}) \\
		\text{ia-enron-employees} & \textrm{0.719} & \textrm{0.759} & \textrm{0.550} & \textbf{0.777}  & (\text{+13.00\%}) \\
		\text{ia-radoslaw-email} & \textrm{0.734} & \textrm{0.741} & \textrm{0.615} & \textbf{0.811} &  (\text{+14.83\%}) \\
		\text{ia-email-EU} & \textrm{0.820} & \textrm{0.860} & \textrm{0.650} & \textbf{0.890} &  (\text{+12.73\%}) \\
		\text{fb-forum} & \textrm{0.670} & \textrm{0.790} & \textrm{0.640} & \textbf{0.826}  &  (\text{+15.25\%}) \\
		\text{soc-bitcoinA} & \textrm{0.840} & \textrm{0.870} & \textrm{0.670} & \textbf{0.891} &  (\text{+10.96\%}) \\
		\text{soc-wiki-elec} & \textrm{0.820} & \textrm{0.840} & \textrm{0.620} & \textbf{0.857} &  (\text{+11.32\%}) \\
		
		\bottomrule
		\multicolumn{6}{l}{\footnotesize $^{\star}$\textsc{Gain} = 
		mean gain in AUC averaged over all embedding methods.} \\
	\end{tabularx}
	\vspace{-2mm}
\end{table}

\subsection{Comparison} \label{sec:comparison}
\noindent
We evaluate the performance of the proposed framework on the temporal link prediction task. 
To generate a set of labeled examples for link prediction, we first sort the edges in each graph by time (ascending) and use the first $75\%$ for representation learning.
The remaining $25\%$ are considered as positive links and we sample an equal number of negative edges randomly. 
Since the temporal network is a multi-graph where an edge between two nodes can appear multiple times with different timestamps, 
we take care to ensure edges that appear in the training set do not appear in the test set.
We perform link prediction on this labeled data $\mathcal{X}$ of positive and negative edges.
After the embeddings are learned for each node, we derive edge embeddings by combining the learned embedding vectors of the corresponding nodes.
More formally, given embedding vectors $\vz_i$ and $\vz_j$ for node $i$ and $j$, 
we derive an edge embedding vector $\vz_{ij} = \Phi(\vz_i, \vz_j)$
where 
\begin{equation} \label{eq:embedding-functions}
\nonumber
\Phi \in \big\lbrace(\vz_i + \vz_j)\big/2,\;\; \vz_i \odot \vz_j,\;\; \abs{\vz_i - \vz_j},\;\; (\vz_i - \vz_j)^{\circ 2}\big\rbrace
\end{equation}\noindent
and $\vz_i \odot \vz_j$ is the element-wise (Hadamard) product and $\vz^{\circ 2}$ is the Hadamard power.
We use logistic regression (LR) with hold-out validation of $25\%$.
Experiments are repeated for 10 random seed initializations and the average performance is reported. 
Unless otherwise mentioned, we use ROC AUC (denoted as AUC for short) to evaluate the models and use the same number of dimensions $D$ for all models.

To compare the methods fairly, we ensure all baseline methods use the same amount of information for learning.
In particular, the number of \emph{temporal context windows} is 
\begin{equation} \label{eq:stopping-criterion}
\beta  = R \times N \times (L - \omega + 1)
\end{equation}\noindent
where $R$ denotes the number of walks for each node and $L$ is the length of a random walk required by the baseline methods. 
Recall that $R$ and $L$ are \emph{not} required by CTDNE and are only used above to ensure that all methods use exactly the same amount of information for evaluation purposes.
Note since CTDNE does not collect a fixed amount of random walks (of a fixed length) for each node as done by many other embedding methods~\cite{deepwalk,node2vec}, instead the user simply specifies the $\#$ of temporal context windows (expected) per node and  the total number of temporal context windows $\beta$ is derived as a multiple of the number of nodes $N=|V|$.
Hence, CTDNE is also easier to use as it requires a lot less hyperparameters that must be carefully tuned by the user.
Observe that it is possible (though unlikely) that a node $u \in V$ is not in a valid temporal walk, \ie, the node does not appear in any temporal walk
$S_t$
with length at least $|S_t| > \omega$.
If such a case occurs, we simply relax the notion of temporal random walk for that node by ensuring the node appears in at least one random walk of sufficient length, even if part of the random walk does not obey time.
As an aside, relaxing the notion of temporal random walks by allowing the walk to sometimes violate the time-constraint can be viewed as a form of regularization.

Results are shown in Table~\ref{table:link-pred-results}.
For this experiment, we use the simplest CTDNE variant from the proposed framework
and did not apply any \emph{additional bias} to the selection strategy. 
In other words, both $\mathbb{F}_s$ and $\mathbb{F}_{\Gamma}$ are set to the uniform distribution. 
We note, however, that since temporal walks are time-obeying (by Definition~\ref{def:temporal-walk}), the selection is already biased towards edges that appear later in time as the random walk traversal does not go back in time. 
In Table~\ref{table:link-pred-results}, the proposed approach is shown to perform consistently better than DeepWalk, node2vec, and LINE. 
This is an indication that important information is lost when temporal information is ignored.
Strikingly, the CTDNE model does not leverage the bias introduced by node2vec~\cite{node2vec}, and yet still outperforms this model by a significant margin.
We could have generalized node2vec in a similar manner using the proposed framework from Section~\ref{sec:framework}. 
Obviously, we can expect to achieve even better predictive performance by using the CTDNE framework to derive a continuous-time node2vec generalization by replacing the notion of random walks in node2vec with the notion of \emph{temporal random walks} biased by the (weighted) distributions $\mathbb{F}_s$ (Section~\ref{sec:selection-of-start-time}) and $\mathbb{F}_{\Gamma}$ (Section~\ref{sec:temporal-random-walk}).

\begin{table}[h!]
\vspace{-4mm}
\centering
\setlength{\tabcolsep}{3.0pt}
\renewcommand{\arraystretch}{1.15}
\small
\fontsize{8}{9}\selectfont
\caption{Results for Different CTDNE Variants}
\label{table:variants-link-pred-results}
\vspace{-2.4mm}
\begin{tabularx}{1.00\linewidth}{ll  c XXXX @{}} 
\multicolumn{7}{p{1.0\linewidth}}{\footnotesize 
$\mathbb{F}_s$ is the distribution for initial edge sampling and $\mathbb{F}_{\Gamma}$ is the distribution for temporal neighbor sampling. 
} \\
\toprule
\multicolumn{2}{c}{\textsc{Variant}} \\
\multicolumn{1}{c}{\fontsize{11}{12}\selectfont $\mathbb{F}_s$} & \multicolumn{1}{c}{\fontsize{11}{12}\selectfont $\mathbb{F}_{\Gamma}$} &&
\multicolumn{1}{l}{\textsf{\fontsize{7.5}{8.5}\selectfont contact}} & \textsf{\fontsize{7.5}{8.5}\selectfont hyper} & \textsf{\fontsize{7.5}{8.5}\selectfont enron} & 
\multicolumn{1}{l}{\textsf{\fontsize{7.5}{8.5}\selectfont rado}} 
\\ \midrule

\fontsize{8.5}{9.5}\selectfont

$\mathbf{Unif}$ (Eq.~\ref{eq:uniform-edge}) & $\mathbf{Unif}$ (Eq.~\ref{eq:uniform-neighbor})  && 
0.913 & 0.671 & 0.777 & 0.811 \\

$\mathbf{Unif}$ (Eq.~\ref{eq:uniform-edge}) & $\mathbf{Lin}$ (Eq.~\ref{eq:linear-penalty})  && 
0.903 & 0.665 & 0.769 & 0.797 \\

$\mathbf{Lin}$ (Eq.~\ref{eq:linear-dist})  & $\mathbf{Unif}$ (Eq.~\ref{eq:uniform-neighbor})  && 
0.915 & 0.675 & 0.773 & 0.818 \\

$\mathbf{Lin}$ (Eq.~\ref{eq:linear-dist})  & $\mathbf{Lin}$ (Eq.~\ref{eq:linear-penalty})  && 
0.903 & 0.667 & 0.782 &  0.806 \\

$\mathbf{Exp}$ (Eq.~\ref{eq:exponential-dist})  & $\mathbf{Exp}$ (Eq.~\ref{eq:exponential-penalty})  && 
\textbf{0.921}	& 0.681	& \textbf{0.800}	& 0.820
\\

$\mathbf{Unif}$ (Eq.~\ref{eq:uniform-edge})  & $\mathbf{Exp}$ (Eq.~\ref{eq:exponential-penalty})  && 
0.913 & 0.670 & 0.759 &  0.803 
\\

$\mathbf{Exp}$ (Eq.~\ref{eq:exponential-dist})  & $\mathbf{Unif}$ (Eq.~\ref{eq:uniform-neighbor})  && 
0.920 &	\textbf{0.718}	& 0.786	& \textbf{0.827}
\\

$\mathbf{Lin}$ (Eq.~\ref{eq:linear-dist})  & $\mathbf{Exp}$ (Eq.~\ref{eq:exponential-penalty})  && 
0.916 &0.681 & 0.782 &  0.823 \\

$\mathbf{Exp}$ (Eq.~\ref{eq:exponential-dist})  & $\mathbf{Lin}$ (Eq.~\ref{eq:linear-penalty})  && 
0.914 & 0.675 & 0.747 &  0.817\\

		\bottomrule
	\end{tabularx}
	\vspace{-2mm}
\end{table}

In all cases, the proposed approach significantly outperforms the other embedding methods across all dynamic networks (Table~\ref{table:link-pred-results}).
The mean gain in AUC averaged over all embedding methods for each dynamic network is shown in Table~\ref{table:link-pred-results}.
Notably, CTDNE achieves an overall gain in AUC of $11.9\%$ across all embedding methods and graphs.
These results indicate that modeling and incorporating the temporal dependencies in graphs is important for learning appropriate and meaningful network representations.
It is also worth noting that many other approaches that leverage random walks can also be generalized using the proposed framework~\cite{struc2vec,ComE,ASNE,dong2017metapath2vec,lee17-Deep-Graph-Attention}, 
as well as any future state-of-the-art embedding method.

\vspace{-2mm}
\subsection{Comparing Variants from CTDNE Framework} \label{sec:exp-variants}
\noindent
We investigate three different approaches for $\mathbb{F}_s$ and $\mathbb{F}_{\Gamma}$ giving rise to nine different CTDNE variants by taking all possible combinations of unbiased and biased distributions discussed in Section~\ref{sec:selection-of-start-time} and Section~\ref{sec:temporal-random-walk}.
In particular, we investigated three different approaches to sample 
(1) the starting temporal edge $e_*$ via $\mathbb{F}_s$, and
(2) each subsequent edge in a temporal random walk via $\mathbb{F}_{\Gamma}$. 
For learning dynamic node embeddings in an online fashion, $\mathbb{F}_s$ is not required since for each new edge $(i,j,t)$ in the graph stream, we sample a number of temporal walks ending at $(i,j)$ and use these to update the embedding.

Overall, we find that using a biased distribution (\eg, linear or exponential) improves predictive performance in terms of AUC when compared to the uniform distribution on many graphs.
For others however, there is no noticeable gain in performance. 
This can likely be attributed to the fact that most of the dynamic networks investigated have a relatively short time span (more than 3 years at most). 
Table~\ref{table:variants-link-pred-results} provides results for a few other variants from the framework.
In particular, Table~\ref{table:variants-link-pred-results} shows the difference in AUC when 
applying a biased distribution to the initial edge selection strategy $\mathbb{F}_s$ as well as the temporal neighbor selection $\mathbb{F}_{\Gamma}$ for the temporal random walk.
Interestingly, using a biased distribution for $\mathbb{F}_s$ seems to improve more on the tested datasets. 
However, for \text{ia-enron-employees}, the best result can be observed when both distributions are biased.

\subsection{Continuous vs. Discrete Approximation-based Embeddings}
\noindent
We also investigate the difference between discrete-time models that learn embeddings from a sequence of discrete snapshot graphs, and the class of continuous-time embeddings proposed in this paper.
\begin{Definition}[\sc DTDN Embedding]
\label{def:DTDNE}
A discrete-time dynamic network embedding (DTDNE) is defined as any embedding derived from a sequence of discrete static snapshot graphs $\mathcal{G} = \{G_1,G_2,\ldots,G_t\}$.
This includes any embedding learned from temporally smoothed static graphs or any representation derived from the initial sequence of discrete static graphs.
\end{Definition}\noindent
Previous work for temporal networks have focused on DTDNE methods as opposed to the class of CTDNE methods proposed in this work.
Notice that DTDNE methods use \emph{approximations} of the actual dynamic network whereas the
CTDN embeddings do not and leverage the actual valid temporal information without any temporal loss.
In this experiment, we create discrete snapshot graphs and learn embeddings for each one using the previous approaches.
As an example, suppose we have a sequence of $T=4$ snapshot graphs where each graph represents a day of activity and further suppose $D=128$.
For each snapshot graph, we learn a $(D/T)$-dimensional embedding and concatenate them all to obtain a $D$-dimensional embedding and then evaluate the embedding for link prediction as described previously.

\begin{table}[b!]
\vspace{-8mm}
\centering
\small
\fontsize{8}{9}\selectfont
\renewcommand{\arraystretch}{1.10} 
\setlength{\tabcolsep}{2.0pt}
\caption{Results Comparing DTDNEs to CTDNEs (AUC)}
\label{table:link-pred-results-discrete-model}
\vspace{-2.4mm}
\begin{tabularx}{1.0\linewidth}{@{}r cc cc c @{}rH}

\multicolumn{8}{p{1.0\linewidth}}{\footnotesize 
CTDNE-Unif uses uniform for both $\mathbb{F}_s$ and $\mathbb{F}_{\Gamma}$ whereas CTDNE-Opt selects the distributions via model learning (and hence corresponds to the best model).
} \\

		\toprule
		\textbf{Dynamic Network} 
		&& \textbf{DTDNE} 
		&& \textbf{CTDNE-Unif} \; 
		& \textbf{CTDNE-Opt} 
		& \;(\textsc{Gain}) 
		\\
		\midrule
	
		\text{ia-contact} && 
		\textrm{0.843} &&
		\text{0.913}  &
		\textbf{0.921} & 
		(\text{+8.30\%}) 
		\\

		\text{ia-hypertext} && 
		0.612 && 
		\text{0.671} & 
		\textbf{0.718} & 
		(\text{+9.64\%}) 
		\\
		
		\text{ia-enron-employees} && 
		0.721 && 
		\text{0.777} & 
		\textbf{0.800} & 
		(\text{+7.76\%}) 
		\\
		
		\text{ia-radoslaw-email} && 
		0.785 && 
		\text{0.811} & 
		\textbf{0.827} & 
		(\text{+3.31\%}) 
		\\
		
		\bottomrule
		\multicolumn{8}{p{1.0\linewidth}}{\footnotesize  $^{\star}$\textsc{Gain} = 
		mean gain in AUC averaged over all embedding methods.} \\
	\end{tabularx}
\end{table}

A challenging problem common with DTDNE methods is how to handle nodes that are not active in a given static snapshot graph $G_i$ (\ie, the node has no edges that occur in $G_i$).
In such situations, we set the node embedding for that static snapshot graph to all zeros.
However, we also investigated using the node embedding from the last active snapshot graph as well as setting the embedding of an inactive node to be the mean embedding of the active nodes in the given snapshot graph and observed similar results.

More importantly, unlike DTDNE methods that have many issues and heuristics required to handle them (\eg, the time-scale, how to handle inactive nodes, etc), CTDNEs do not.
CTDNEs also avoid many other issues~\cite{CTDNE} discussed previously that arise from DTDN embedding methods that use a sequence of discrete static snapshot graphs to approximate the actual dynamic network.
For instance, it is challenging and unclear how to select the ``best'' most appropriate time-scale used to create the sequence of static snapshot graphs; and the actual time-scale is highly dependent on the temporal characteristics of the network and the underlying application.
More importantly, all DTDNs (irregardless of the time-scale) are \emph{approximations} of the actual dynamic network.
Thus, any DTDN embedding method is inherently lossy 
and is only as good as the discrete approximation of the CTDN (graph stream).
Results are provided in Table~\ref{table:link-pred-results-discrete-model}. 
Since node2vec always performs the best among the baseline methods (Table~\ref{table:link-pred-results}),
we use it as a basis for the DTDN embeddings.
For brevity, we show results for each of the networks used previously in Table~\ref{table:variants-link-pred-results}.
Overall, the proposed CTDNEs perform better than DTDNEs as shown in Table~\ref{table:link-pred-results-discrete-model}.
Note that CTDNE in Table~\ref{table:link-pred-results-discrete-model} corresponds to using uniform for both $\mathbb{F}_s$ and $\mathbb{F}_{\Gamma}$.
Obviously, better results can be achieved by learning $\mathbb{F}_s$ and $\mathbb{F}_{\Gamma}$ automatically as shown in Table~\ref{table:variants-link-pred-results}.
The gain in AUC for each graph is shown in the rightmost column in Table~\ref{table:link-pred-results-discrete-model}.
The mean gain in AUC of CTDNE compared to DTDNE over all graphs is $7.25\%$.

\definecolor{typeTwoColor}{RGB}{222,45,38} 
\definecolor{typeOneColor}{RGB}{49,130,189} 
\definecolor{typeThreeColor}{RGB}{77,172,38} 

\makeatletter
\global\let\tikz@ensure@dollar@catcode=\relax
\makeatother
\tikzstyle{every node}=[font=\large,line width=1.5pt]
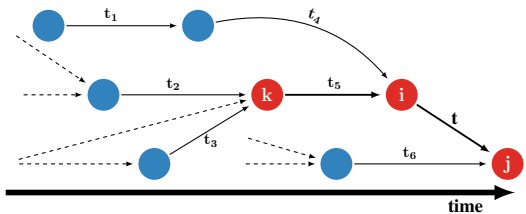
\begin{figure}[h!]
\centering
\begin{center}

\scalebox{0.5}{
\centering
\begin{tikzpicture}[->,>=latex,shorten >=2.4pt,auto,node distance=2.6cm,thick,
main node/.style={circle,draw=white,fill=typeOneColor,draw,text=white,minimum width=0.9cm,font=\sffamily\Large\bfseries},
red node/.style={circle,draw=white,fill=typeTwoColor,draw,text=white,minimum width=0.9cm,font=\sffamily\Large\bfseries},
white node/.style={circle,draw=white,fill=white,text=white,draw,text=white,minimum width=0.9cm,font=\sffamily\Large\bfseries},
whitesmall node/.style={circle,fill=white,draw=white,minimum width=0.02cm,font=\sffamily\Large\bfseries}]

\node[main node] (3) {}; 
\node[main node] (10) [left of=3, left=9mm] {}; 
\node[main node] (1) [below left of=3, left=2mm] {}; 
\node[main node] (4) [below right of=1, left=0.1mm] {}; 
\node[white node] (44) [below left of=1, left=4mm] {}; 
\node[white node] (444) [below left of=1, left=8mm, above=1mm] {}; 
\node[white node] (55) [above left of=1, left=4mm] {}; 
\node[white node] (66) [left of=1] {}; 
\node[red node] (2) [below right of=3] {$\mathbf{k}$}; 
\node[main node] (9) [below right of=2] {}; 
\node[red node] (5) [right of=2, right=5mm] {$\mathbf{i}$};
\node[red node] (6) [below right of=5, right=5mm] {$\mathbf{j}$};

\node[white node] (88) [above of=2, below=13mm, left=0mm] {}; 
\node[whitesmall node] (99) [below of=2, above=15mm, left=5mm] {};
\node[whitesmall node] (999) [left of=9] {};

\node[white node] (7) [left of=1] {$\mathbf{---}$}; 
\node[white node] (8) [right of=5] {$\mathbf{---}$}; 
\node[white node] (111) [below of=66, left=0mm] {}; 
\node[white node] (222) [below of=8, right=5mm] {}; 

\tikzstyle{LabelStyle}=[below=3pt]
\path[every node/.style={font=\large \sffamily}] 
(10) edge [left] node [above left] {$\mathbf{t_1}$} (3) 
(1) edge [right] node [above left] {$\mathbf{t_2}$} (2) 
	(4)  edge [] node[anchor=center,below] {$\mathbf{t_3}$} (2)
	(9)  edge [right] node[above left] {$\mathbf{t_6}$} (6)
	(55) edge [dashed, left] node[below left] {} (1)
(44) edge [dashed, left] node[below left] {} (4)
(44) edge [dashed, right] node[below right] {} (2)
	(66) edge [dashed, left] node[below left] {} (1)
(99) edge [dashed] node[below=0pt] {} (9)
(999) edge [dashed] node[below=0pt] {} (9)

	(3) edge[bend left]  node[sloped,anchor=center,above] {$\mathbf{t_4}$} (5) 
	(5) edge [line width=0.5mm]  node[anchor=center,above] {\Large \sffamily \bf t} (6) 
	(2) edge[right,  line width=0.5mm]  node[sloped,anchor=center,above] {$\mathbf{t_5}$} (5) 

(111) edge [thick,line width=1.5mm,draw=black, below right] node [below right] {\;\quad\quad\quad\quad\quad\quad\quad\quad\quad\quad\quad \Large \bf time} (222);
\end{tikzpicture}
}
\end{center}

\vspace{-5mm}
\caption{
Temporal Walks for Online CTDNEs.
Given a new edge $(i, j,t)$ at time $t$, 
we immediately add it to the graph and then
sample temporal walks ending at edge $(i,j)$ 
and use them to update the relevant embeddings.
An example of a temporal walk is $k \!\! \rightarrow \! i \!\! \rightarrow \! j$ (red nodes).
Note $t > t_6 > t_5 > t_4 > t_3 > t_2 > t_1$.
In this example, $k$ and $j$ are the training instances.
Hence, $\vz_i$ is updated every time $i$ is used in a temporal edge.
}
\label{fig:online-temporal-walk}
\vspace{-5mm}
\end{figure}

\subsection{Incremental Learning of Node Embeddings} \label{sec:exp-online-learning}
\noindent
For some applications, it is important to incrementally learn and update embeddings from edges as soon as they arrive in a streaming fashion.
In such a streaming online setting, we perform fast partial updates to obtain updated embeddings in real-time.
Given an edge $(i,j,t)$ at time $t$, we simply obtain a few temporal walks ending at $(i,j)$ and use these to obtain the updated embeddings.
An example is shown in Figure~\ref{fig:online-temporal-walk}.
In these experiments, we use online SGD updates (online word2vec)~\cite{kaji2017incremental,peng2017incrementally,luo2015online,li2017psdvec} to incrementally learn the embeddings as new edges arrive.
However, other incremental optimization schemes can be used as well (\eg, see~\cite{duchi2011adaptive,flaxman2005online,zhao2012fast,schraudolph2007stochastic,ge2015escaping,ying2008online}).
We vary the number of temporal walks sampled for every new edge that arrives.
Results are shown in Table~\ref{table:streaming-results}.
Notably, it takes on average only a few milliseconds to update the embeddings across a wide variety of temporal network streams.
These results are from a python implementation of the approach and thus the runtime to process a single edge in the stream can be significantly reduced even further using a C++ implementation of the incremental/online learning approach.

\begin{table}[h!]
\centering
\renewcommand{\arraystretch}{1.15}
\scriptsize
\footnotesize
\renewcommand{\arraystretch}{1.10} 
\setlength{\tabcolsep}{6.0pt}
\caption{Streaming Online Network Embedding Results}
\vspace{-2.4mm}
\label{table:streaming-results}
\begin{tabular}{r HH llH HH H ccc HH}
\multicolumn{12}{p{0.9\linewidth}}{\footnotesize 
Average runtime (in milliseconds) per edge is reported.
We vary the number of walks per new edge from 1 to 10.
Recall $|E_T|$ = \# of \emph{temporal edges} and $\bar{d}$ = average temporal node degree. 
} \\
\toprule
& & & & & & & & & \multicolumn{3}{c}{\bf $\mathbf{Time}$ (ms.)} \\

\cmidrule(l{3pt}r{3pt}){6-12}

\textbf{Dynamic Network} & & & $|E_T|$ & $\bar{d}$ & & & & &  
$\mathbf{1}$ &
$\mathbf{5}$ &
$\mathbf{10}$ &
\\
\midrule

\text{ia-hypertext} & & & 20.8K & 368.5 & && &&
2.769 & 3.721 & 4.927 \\

\text{fb-forum} & & & 33.7K & 75.0 & && &&
2.875 & 3.412 & 4.230 \\

\text{soc-wiki-elec} & & & 107K & 30.1 & && &&
2.788 & 3.182 & 3.813 \\

\text{ia-contact} & & & 28.2K & 206.2 & && && 
2.968 & 4.490 & 6.119 \\

\text{ia-radoslaw-email} & & & 82.9K & 993.1 & && &&
3.266 & 5.797 & 8.916 \\

\text{soc-bitcoinA} & & & 24.1K & 12.8 & && &&
2.679 & 2.965 & 3.347 \\

\bottomrule
\end{tabular}
\vspace{-2mm}
\end{table}

\subsection{Discussion} \label{sec:exp-discussion}
\noindent
Recently, there has been a wide variety of works that are based on the key idea proposed in our shorter manuscript from early 2018~\cite{CTDNE}, which is to leverage temporal walks to extend existing embedding methods, \eg, see~\cite{huang2020temporal,node2bits-arxiv,kumar2019predicting,beres2019node,trivedi2018dyrep,sajjad2019efficient,heidari2020evolving}.
This includes temporal walks based on either BFS and/or DFS.
For temporal clarity, these works were not compared against or reviewed previously in detail.
However, we briefly summarize some of these recent works.
In particular, node2bits~\cite{node2bits-arxiv} used the idea of temporal walks to learn space-efficient dynamic embeddings for user stitching.
There has been some work for temporal bipartite edge streams where an RNN-based model is proposed to embed users and items by leveraging the notion of a 1-hop temporal walk used in this work~\cite{kumar2019predicting}.
Other work has used the proposed temporal walks to learn embeddings for tracking and measuring node similarity in edge streams~\cite{beres2019node}.
More recently, some work has also used the proposed idea of leveraging temporal walks for embeddings to extend Graph Neural Networks (GNNs)~\cite{huang2020temporal}.
In particular, these works use BFS-based temporal walks.
Notably, all of these works are based on complex deep learning techniques that leverage temporal walks, yet they achieve comparable results on some problems.

\section{Challenges \& Future Directions} \label{sec:discussion}

\noindent\textbf{Attributed Networks \& Inductive Learning}:
The proposed framework for learning \emph{dynamic node embeddings}
can be easily generalized to \emph{attributed networks} and for \emph{inductive learning} tasks in temporal networks (graph streams) using the ideas introduced in~\cite{role2vec,ahmed17Gen-Deep-Graph-Learning}.
More formally, the notion of attributed/feature-based walk (proposed in~\cite{role2vec,ahmed17Gen-Deep-Graph-Learning}) can be combined with the notion of temporal random walk as follows:
\begin{Definition}[\sc Attributed Temporal Walk] 
\label{def:attr-temporal-random-walk}
Let $\vx_i$ be a $d$-dimensional feature vector for node $v_i$.
An attributed temporal walk $S$ of length $L$ is defined as a sequence of adjacent node feature-values 
$\phi(\vx_{i_{1}}), \phi(\vx_{i_{2}}),\ldots, \phi(\vx_{i_{L+1}})$
associated with a sequence of indices $i_{1}, i_{2}, \ldots, i_{L+1}$ such that 
{\smallskip
\begin{compactenum}
\item $(v_{i_{t}}, v_{i_{t+1}}) \in E_T$ for all $1 \leq t \leq L$
\item $\mathcal{T}(v_{i_{t}}, v_{i_{t+1}}) \leq \mathcal{T}(v_{i_{t+1}}, v_{i_{t+2}})$ for $1 \leq t < L$ 
\item $\phi : \vx \rightarrow y$ is a function that maps the input vector $\vx$ of a node to a corresponding feature-value $\phi(\vx)$.
\end{compactenum}\noindent
}\noindent
The feature sequence $\phi(\vx_{i_{1}}), \phi(\vx_{i_{2}}),\ldots, \phi(\vx_{i_{L+1}})$ represents the feature-values that occur during a temporally valid walk, i.e., a walk they obeys the direction of time defined in (2).
\end{Definition}\noindent
Attributed temporal random walks can be either uniform (unbiased) or non-uniform (biased).
Furthermore, the features used in attributed walks can be (i) intrinsic input attributes (such as profession, political affiliation), (ii) structural features derived from the graph topology (degree, triangles, etc; or even node embeddings from an arbitrary method), or both.
Temporal attriuted walks can be sampled for every feature as done in~\cite{node2bits-arxiv}.
In this case, $\phi : \mathbb{R}^{d} \rightarrow \mathbb{R}^{d}$ and thus we have $d$ different feature-based walks for every temporal walk sampled.
Suppose $\phi$ is the identity function, then for an arbitrary temporal walk $\lbrace(v_{i_{1}}, v_{i_{2}}, t_{i_{1}})$, $(v_{i_{2}},v_{i_{3}}, t_{i_{2}}), \ldots, (v_{i_{L}},$ $v_{i_{L+1}}, t_{i_{L}})\rbrace$ such that $t_{i_{1}} \leq t_{i_{2}} \leq \ldots \leq t_{i_{L}}$ we have the following $d$ attributed temporal walks (one per feature):
\begin{align}
\begin{matrix} 
X_{i_{1},1} & X_{i_{2},1} & \cdots & X_{i_{k},1} & \cdots \\
X_{i_{1},2} & X_{i_{2},2} & \cdots & X_{i_{k},2} & \cdots \\
\vdots & \vdots  & \vdots  & \vdots  & \vdots \\
X_{i_{1},d} & X_{i_{2},d} & \cdots & X_{i_{k},d} & \cdots \\
\end{matrix}
\end{align}
A recent work called node2bits~\cite{node2bits-arxiv} leveraged this idea for learning inductive dynamic node embeddings and demonstrated its effectiveness compared to a variety of state-of-the-art methods.
We refer the reader to~\cite{node2bits-arxiv} for detailed results and findings.

\medskip\noindent\textbf{Other Types of Temporal Networks}:
While this work naturally supports temporal networks and graph streams in general, there are many other networks with more specialized characteristics.
For instance, some temporal networks (graph streams) contain edges with start and end times.
Developing CTDNE methods for such temporal networks remains a challenge.
Furthermore, another open and challenging problem that remains to be addressed is how to develop graph stream embedding techniques that require a fixed amount of space.
Other applications may require dynamic node embedding methods that are space-efficient (\eg, by learning a sparse vector representation for each node).

\medskip\noindent\textbf{Temporal Weighting and Bias}:
This paper explored a number of temporal weighting and bias functions for decaying the weights of data that appears further in the past.
More research is needed to fully understand the impact and to understand the types of temporal networks and characteristics that each should be used.
Some early work has focused on temporally weighting the links, nodes, and attributes prior to learning embeddings~\cite{rossi2012dynamic-srl}.
However, this idea has yet to be explored for learning general node embeddings and should be investigated in future work.
Other research should investigate new temporal weighting schemes for links, nodes, and attributes~\cite{rossi2012dynamic-srl}.
Furthermore, one can also incorporate a decay function for each temporal walk such that more temporal influence is given to recent nodes in the walk than to nodes in the distant past. 
Hence, each temporal walk is assigned a sequence of weights which can be incorporated into the Skip-Gram approach.
For instance, in the case of an exponential decay function $\alpha^{t-1} \cdot \alpha^{t-2} \cdots \alpha^{t-k}$.
However, there are many other ways to temporally weight or bias the walk and it is unclear when one approach works better than another.
Future work should systematically investigate different approaches.

\section{Conclusion} \label{sec:conc}
\noindent
In this work, we described a new class of embeddings based on the notion of temporal walks.
This new class of embeddings are learned directly from the temporal network (graph stream) without having to approximate the edge stream as a sequence of discrete static snapshot graphs.
As such these embeddings can be learned in an online fashion as they are naturally amenable to graph streams and incremental updates.
We investigated a framework for learning such dynamic node embeddings using the notion of temporal walks.
The proposed approach can be used as a basis for generalizing existing (or future state-of-the-art) random walk-based embedding methods for learning of dynamic node embeddings from dynamic networks (graph streams).
The result is a more appropriate dynamic node embedding
that captures the important temporal properties of the node in the continuous-time dynamic network.
By learning dynamic node embeddings based on temporal walks, we avoid the issues and information loss that arise when time is ignored or approximated using a sequence of discrete static snapshot graphs.
In contrast to previous work, the proposed class of embeddings are learned from temporally valid information.
The experiments demonstrated the effectiveness of this new class of dynamic embeddings on several real-world networks.

\makeatletter
\IEEEtriggercmd{\reset@font\normalfont\fontsize{7.9pt}{8.40pt}\selectfont}
\makeatother
\IEEEtriggeratref{1}
\bibliography{rossi}  
\bibliographystyle{IEEEtran}

\begin{IEEEbiography}[{
\includegraphics[width=1in,height=1.25in,clip,keepaspectratio]{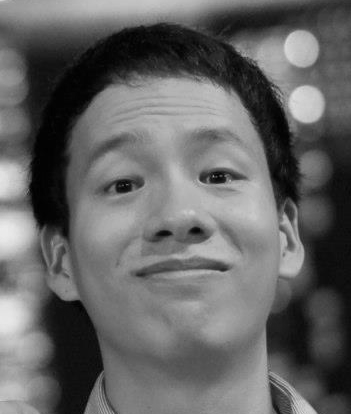}
}]
{John Boaz Lee} is a Research Scientist at Facebook Research. 
He received his Ph.D. from the Computer Science Department at Worcester Polytechnic Institute (WPI) in 2019, and his M.S. in Computer Science from the University of the Philippines - Diliman. 
Dr. Lee's research lies in graph-based deep learning, attention models, and social network analysis.
\end{IEEEbiography}

\vskip -3.0\baselineskip plus -1fil

\begin{IEEEbiography}[{
\includegraphics[width=1in,height=1.25in,clip,keepaspectratio]{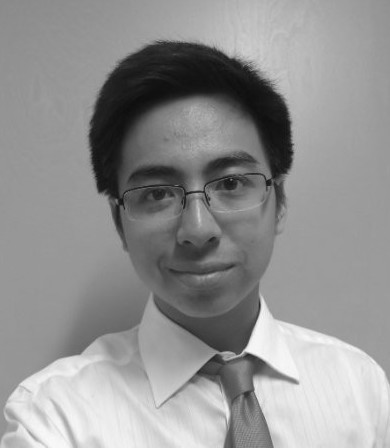}
}]
{Giang Nguyen} is a student at Worcester Polytechnic Institute (WPI). 
His research interests include deep learning and graph mining.
Giang earned his M.S. in Computer Science from Worcester Polytechnic Institute.
\end{IEEEbiography}

\vskip -1.4\baselineskip plus -1fil

\begin{IEEEbiography}[{\includegraphics[width=1in,height=1.25in,clip,keepaspectratio]{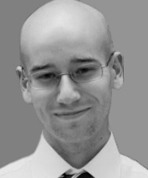}}]
{Ryan A. Rossi} is a research scientist at Adobe Research. 
His research lies in the fields of graph-based machine learning; and spans theory, algorithms, and applications of large complex relational (network/graph) data.
Before joining Adobe Research, he had the opportunity to work at a number of industrial, government, and academic research labs including Palo Alto Research Center (Xerox PARC), Lawrence Livermore National Laboratory (LLNL), Naval Research Laboratory (NRL), NASA Jet Propulsion Laboratory (JPL)/California Institute of Technology, University of Massachusetts Amherst, among others. 
He earned his Ph.D. and M.S. in Computer Science at Purdue University. 
Dr. Rossi was a recipient of the National Science Foundation Graduate Research Fellowship (NSF GRFP), National Defense Science and Engineering Graduate Fellowship (NDSEG), the Purdue Frederick N. Andrews Fellowship, and Bilsland Fellowship.
\end{IEEEbiography}

\vskip -1.4\baselineskip plus -1fil

\begin{IEEEbiography}[{\includegraphics[width=1in,height=1.25in,clip,keepaspectratio]{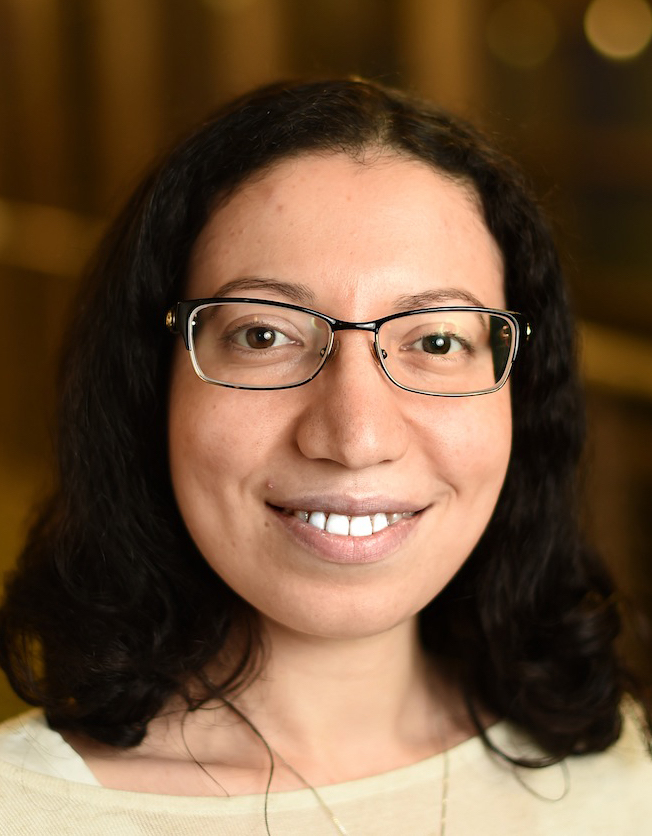}}]
{Nesreen K. Ahmed} is a senior research scientist at Intel Labs. She received her Ph.D. from the Computer Science Department at Purdue University in 2015, and her M.S. in statistics and computer science from Purdue University in 2014.  
In 2018, she is the PC Chair of the IEEE Big Data Conference. 
Dr. Ahmed was a visiting researcher at Facebook, Adobe research, Technicolor, and Intel analytics. 
Her research lies in the field of large-scale machine learning and spans the theory and algorithms of graphs, statistical machine learning methods, and their applications in social and information networks. 
Her research was selected among the best papers of ICDM in 2015, BigMine in 2012, and covered by popular press such as the MIT Technology Review. 
She was selected by UC Berkeley among the top female Rising Stars in computer science and engineering in 2014.
Dr. Ahmed holds 2 U.S. patents and is co-founder of the first interactive network data repository
(http://networkrepository.com).
\end{IEEEbiography}

\vskip -1.4\baselineskip plus -1fil

\begin{IEEEbiography}[{\includegraphics[width=1in,height=1.25in,clip,keepaspectratio]{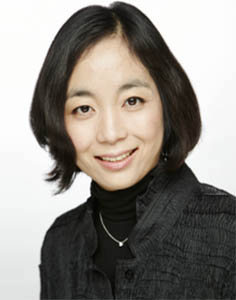}
}]
{Eunyee Koh} is a researcher at Adobe Research. 
She obtained her PhD at the Computer Science Department of Texas A\&M University. 
She graduated in 2002 from the Seoul National University with a bachelor’s in computer science and engineering, and worked for Motorola and Accenture during 2002-2003.
She has been a member of Adobe’s ATL since 2007. 
Her current research interests include semantic analysis and metadata extraction from media, and how to visualize those extracted metadata interactively.
\end{IEEEbiography}

\vskip -1.4\baselineskip plus -1fil

\begin{IEEEbiography}[{\includegraphics[width=1in,height=1.25in,clip,keepaspectratio]{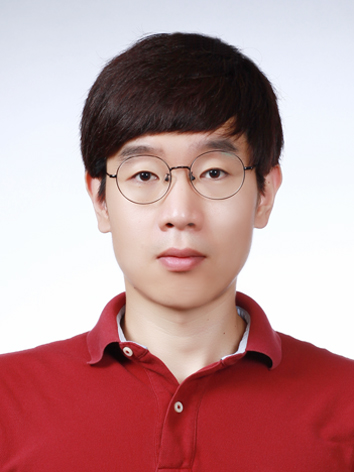}
}]
{Sungchul Kim} is a researchscientist at the Systems Technology Lab of Adobe Research. 
He received his Ph.D. in Computer Science and Engineering at Pohang University of Science and Technology (POSTECH) in 2015.
His research focuses on user behavior modeling and visitor stitching and include representing and analyzing data by using data-mining techniques including regression, ranking, and dimensionality reduction.
Dr. Kim also has worked on recommender system, online advertisement and named entity recognition.
\end{IEEEbiography}

\end{document}